\documentclass[lettersize,journal]{IEEEtran}
\usepackage{amsmath,amsfonts}
\usepackage{algorithmic}
\usepackage{array}
\usepackage[caption=false,font=normalsize,labelfont=sf,textfont=sf]{subfig}
\usepackage{textcomp}
\usepackage{stfloats}
\usepackage{url}
\usepackage{verbatim}
\usepackage{graphicx}

\usepackage{multirow}
\usepackage{multicol}
\usepackage{tabularx}
\usepackage{booktabs}
\usepackage[linesnumbered,ruled,vlined]{algorithm2e}
\usepackage{pifont}
\usepackage[table]{xcolor}
\usepackage{orcidlink}
\usepackage{xpatch}

\usepackage[style=ieee,backref=false,doi=false,isbn=false,url=false,eprint=false,sorting=none, maxnames=6, minbibnames=1]{biblatex}

\allowdisplaybreaks[4]

\usepackage{hyperref}
\hypersetup{
colorlinks=true,
linkcolor=black,
citecolor=blue
}

\def\@IEEEBIOskipN{1.5 \baselineskip}
\xpatchcmd{\IEEEbiography} {\vskip \@IEEEBIOskipN plus 1fil minus 0\baselineskip} {\vskip \@IEEEBIOskipN} {} {\PackageWarning{IEEEBiographyPatch} {Failed to patch IEEEbiography}}
\xpatchcmd{\IEEEbiographynophoto} {\vskip 4\baselineskip plus 1fil minus 0\baselineskip} {\vskip \@IEEEBIOskipN} {} {\PackageWarning{IEEEBiographyPatch} {Failed to patch IEEEbiographynophoto}}
\makeatother

\begin{document}

\title{\textsc{Miles}: Metric Learning with Expandable Subspace \\ for Pre-Trained Model-Based Class-Incremental Learning}

\author{Kai Jiang$^{\orcidlink{0009-0002-7406-9177}}$, Zisong Lin, Hongyuan Zhang$^{\orcidlink{0000-0003-4274-7332}}$, Xueru Bai$^{\orcidlink{0000-0001-9283-1810}}$,~\IEEEmembership{Senior Member,~IEEE}, and Xuelong Li$^{\orcidlink{0000-0003-2924-946X}}$,~\IEEEmembership{Fellow,~IEEE}
\thanks{This work was supported in part by the National Natural Science Foundation of China under Grant 62425113 and Grant 62131020. (\textit{Corresponding author}:~Xueru Bai).}
\thanks{Kai Jiang, Zisong Lin and Xueru Bai are with the National Key Laboratory of Radar Signal Processing, Xidian University, Xi’an 710071, China (e-mail: xrbai@xidian.edu.cn).}
\thanks{Hongyuan Zhang is with The University of Hong Kong, Hong Kong SAR, China (e-mail: hyzhang98@gmail.com).}
\thanks{Xuelong Li is with the Institute of Artificial Intelligence (TeleAI) of China Telecom, China (e-mail: xuelong\_li@ieee.org).}
}

\maketitle

\begin{abstract}
Class Incremental Learning~(CIL) aims to learn new concepts consistently from a data stream without forgetting. Unlike typical CIL methods which need to learn a model from scratch, pre-trained model~(PTM) can easily adapt to a new task with fine-tuning. However, existing PTM-based CIL methods fail to achieve a trade-off between performance and computational expenditure, i.e., they either adopt the same parameter space so that leading catastrophic forgetting, or expand a new branch for each task but adding more computational cost. To this end, we propose MetrIc Learning with Expandable Subspace~(\textsc{Miles}) to harness the prior information within pre-trained knowledge, thereby orchestrating an efficient expansion of the parameter space through guided optimization. Specifically, it decouples the learnable modules with the pre-trained model, exploiting prior information from intermediate features of the backbone network to enable more flexible parameter expansion. Then, a central loss is adopted to guide the new category to cluster towards the corresponding prototype in the new task subspace while incorporating an auxiliary distance regularization term to maintain metric equilibrium across tasks. Extensive experiments on six benchmark datasets demonstrate that \textsc{Miles} achieves state-of-the-art performance in various CIL settings.
\end{abstract}


\section{Introduction}
\IEEEPARstart{D}{eep} learning has demonstrated remarkable success in the field of computer vision and has been gradually extended to real-world applications~\cite{application1, application2, RN, application3}. However, existing AI systems cannot continuously learn new concepts without forgetting, which hinders their rapid development in many fields~\cite{Catastrophic01, Catastrophic02, UNIFIER}. For instance, autonomous driving requires AI systems to adapt to seasonal changes~\cite{auto-driving-TIP}, and medical systems require AI assistance continually learn to recognize new diseases~\cite{medical-TIP}. Therefore, continual learning is proposed to solve these real-world requirements~\cite{continual01, continual02, continual03}.

As an important direction of continual learning, class incremental learning~(CIL) requires models to learn new categories sequentially without forgetting old ones~\cite{TIPCIL1, TIPCIL2, ViewMask}. Conventional CIL methods can be roughly divided into three types, i.e., regularization-based methods~\cite{lwf, ewc, yu2020semantic}, rehearsal-based methods~\cite{icarl, rmm, luo2023class}, and architecture-based methods~\cite{der, foster, beef, memo}. Since these methods learn a model from the scratch, they usually lead to a long training time and poor generalization capability.

With the advancement of pre-training techniques, the approach of pre-training on upstream tasks followed by fine-tuning on downstream tasks has been widely adopted in real-world applications~\cite{chen2021large, Pinoise01, chen2022learning}, and has also shed lights on developing CIL methods. By employing pre-trained weights to initialize models, training speed and generalization capacity can experience profound elevation~\cite{aper}. Although the pre-trained model~(PTM) furnishes a wealth of prior knowledge, directly fine-tuning it across multiple downstream tasks still encounters severe catastrophic forgetting~\cite{Catastrophic01, Catastrophic02}. To resolve this challenge, pre-trained model-based incremental learning (PTM-based CIL) is proposed to exploit the prior information learned from a large scale of datasets, such as some low-level vision characteristics like edges or textural patterns, so that enhancing the generalization ability in downstream tasks~\cite{zhou2024continual}. 

\begin{figure}[t]
    \centering
    \includegraphics[width=3in]{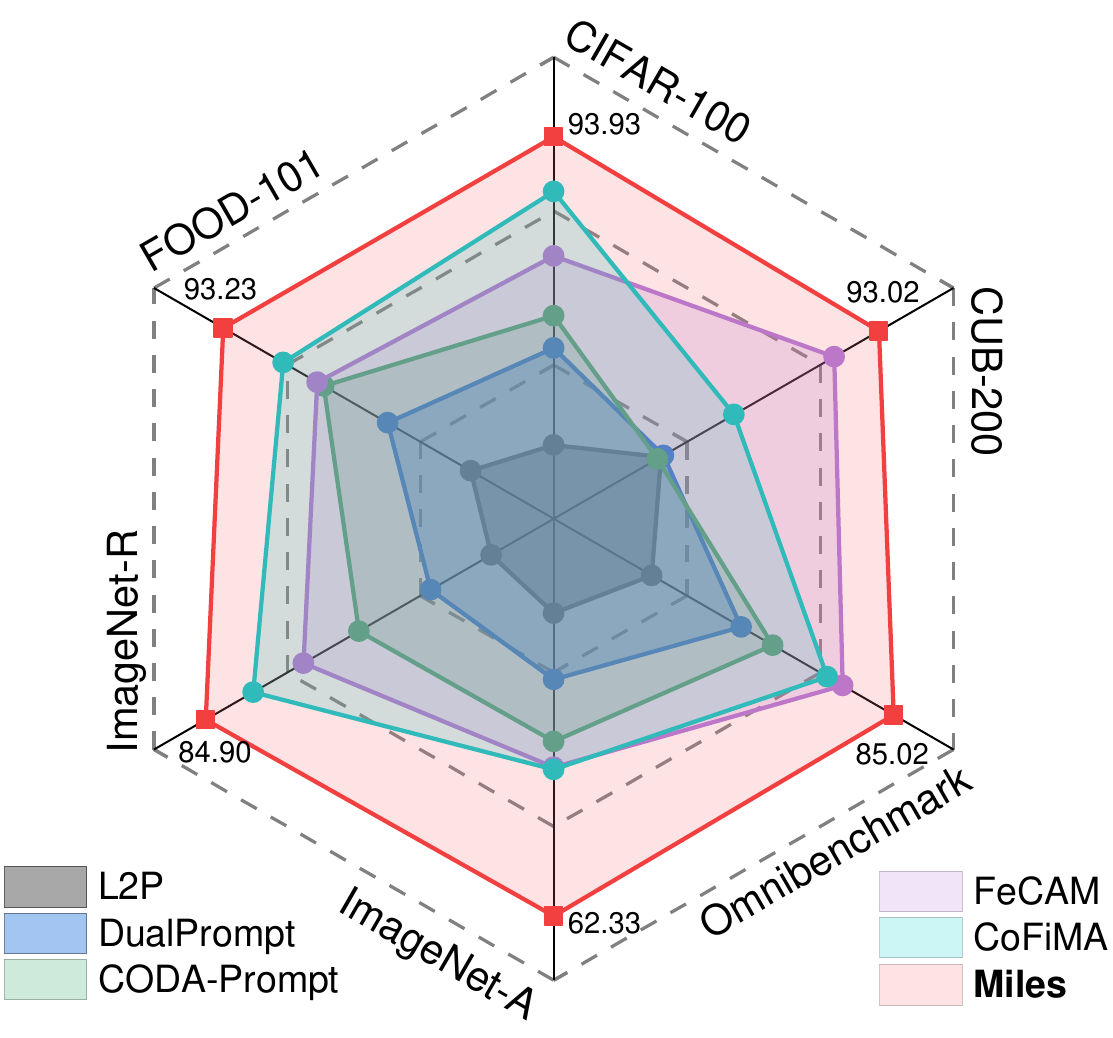}
    \caption{Average accuracies of incremental trends for different approaches.}
	\label{fig:sota}
\end{figure}

\begin{figure}[t]
    \centering
    \includegraphics[width=3.4in]{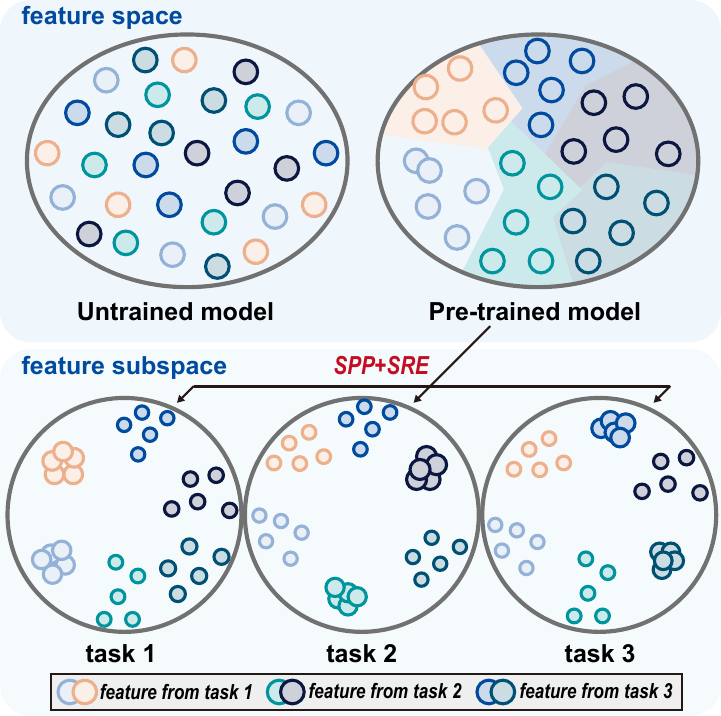}
    \caption{Motivation of the \textsc{Miles}.}
	\label{fig:motivation}
\end{figure}

Since we observe that unseen categories during pre-training usually exhibit clustering rather than dispersion in the feature space of the PTM, as illustrated in Fig.~\ref{fig:motivation}, we are motivated to integrating metric learning with pre-trained models in CIL~\cite{li2023deep}. By initializing the model with pre-trained weights, a reliable prototype~\cite{snell2017prototypical} can be computed without training for each category of downstream tasks. Then, we aim to adopt these prototypes as prior knowledge from pre-training to guide the optimization of adapting new task instead of only applying the current task data, thus avoiding model preference for new tasks. Specifically, we propose \textit{Subspace Prototype Projection}~(\textbf{SPP}) to cascade a tiny learnable module after the backbone for each task to reduce the distance between features and the corresponding prototypes. In theory, features after the mapping process should be close to the prototype of the ground truth and be away from the out-of-task prototypes.

However, the cascaded learnable modules following the backbone network usually exhibit constrained representational capacity. Some existing methods prefer to embed the trainable modules into the pre-trained backbone to strengthen model representation~\cite{lora, Pinoise03, vpt, adaptformer}. But these approaches present two limitations: primarily, they induce modifications to pretrained backbone networks, thereby complicating architectural substitutions; secondly, their requirement for multiple forward propagation induces inference time increases linearly with the task quantity~\cite{ease}.

Therefore, we attempt to devise a CIL method which can adapt new tasks with minimal modification in the pre-trained network and propagates forward only once. Based on this idea, we propose \textit{Subspace Representation Extension}~(\textbf{SRE}), which decouples the pre-trained weights with learnable weights and extends a low-rank sub-network for each task to learn all the intermediate features from the pre-trained model, thereby enhancing the model representational capacity without modification of the pre-trained network. Moreover, employing metric learning to learn subsequent tasks can achieve CIL naturally, which only need to compute prototypes of new task categories and optimize features towards these prototypes. However, the model easily overfits on these downstream tasks, which is manifested as the average distance on a certain task being significantly smaller than the average distance on previous tasks. In view of this, we propose \textit{Distance Regularization}~(\textbf{DR}), which adds a distance regularization term to maintain a consistent average distance for each task. The main contributions of our work are three-folds:

\begin{itemize}
    \item We integrate metric learning and pre-trained models into CIL and propose \textit{Subspace Prototype Projection}~(SPP) to optimize the distance between current task features and corresponding prototypes in the new task subspace. 
    \item We propose \textit{Subspace Representation Extension}~(SRE) to add a tiny sub-network to learn the intermediate features from PTM, aiming to strengthen the model representational capacity with minimal modification of pre-trained network.
    \item We propose \textit{Distance Regularization}~(DR) to add a distance regularization term to maintain a consistent average distance for each task, with the aim of solving the overfitting issue for new tasks.
\end{itemize}

As shown in Fig.~\ref{fig:sota}, \textsc{Miles} shows state-of-the-art performance in multiple datasets.

\section{Related works}
In this section, we briefly introduce some related works including conventional class incremental learning and PTM-based class-incremental learning:
\paragraph{Class incremental learning (CIL)} aims to learn a unified model from a task data stream containing multiple disjoint categories. Conventional CIL approaches include three types. The first type is regularization-based methods, which aim to penalize modifications of crucial parameters~\cite{lwf, podnet, cscct, 10605121, TIPCIL3, mtd} by adding a regularization term to the optimization function~\cite{shi2026protoconnet, qi2025classwise}. The second type is rehearsal-based methods, which construct an exemplar set to save a few samples from previous tasks for future training~\cite{icarl, rmm, bang2021rainbow, TIPCIL4,luo2023class}. The last type is architecture-based methods, which aim to increase model capacity by adding learnable parameters for a new task to solve the plasticity-stability dilemma~\cite{der, dytox, lyu2023overcoming}. However, all three methods need to train a model from scratch, which leads to a longer training time and weaker generalization ability.

\begin{figure*}[t]
    \centering
    \includegraphics[width=6.8in]{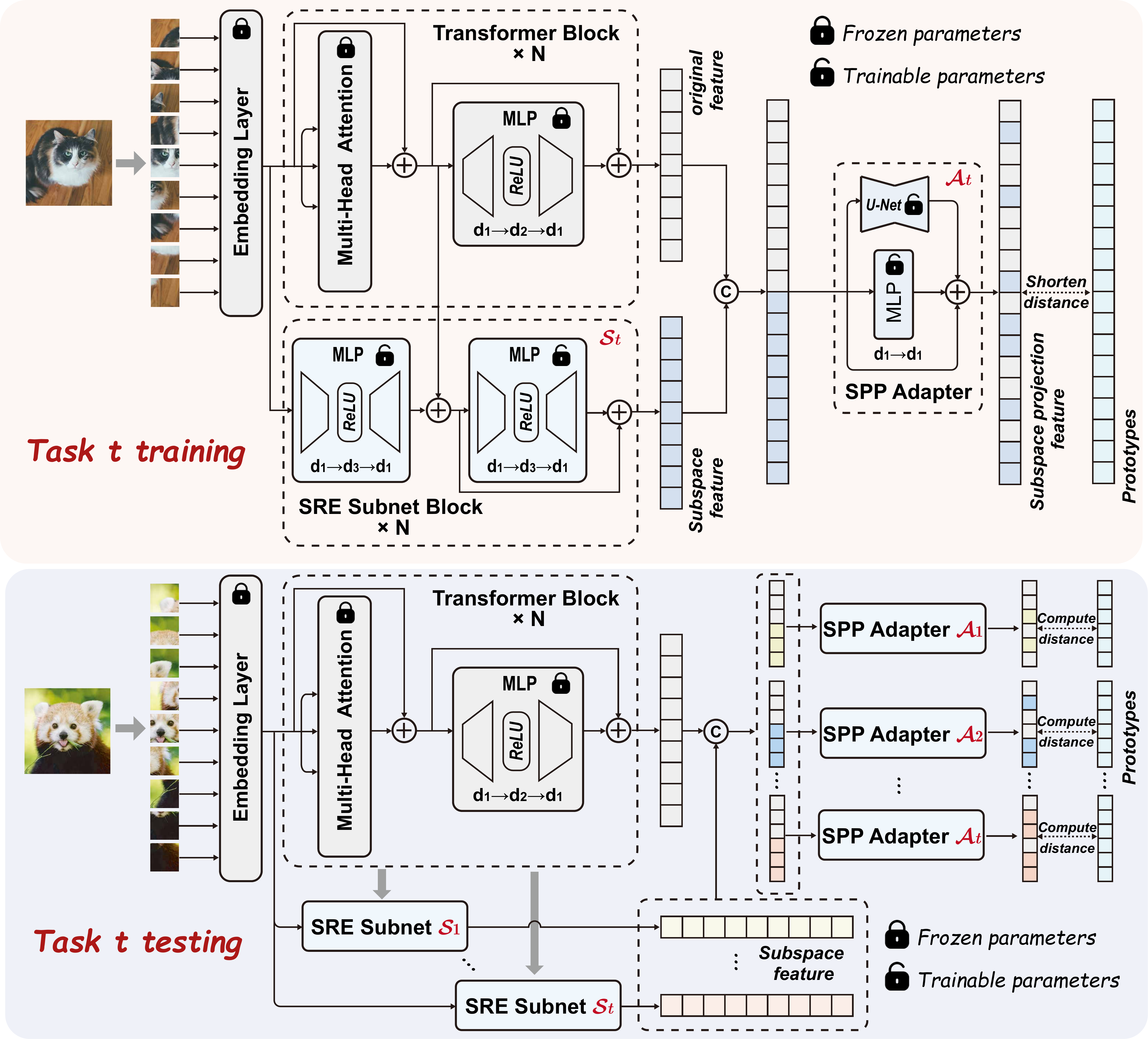}
    \caption{Overall structure of the proposed method.}
	\label{fig:overall}
\end{figure*}

\noindent
\paragraph{Pre-trained model-based class incremental learning (PTM-based CIL)} aims to utilize pre-trained models to obtain a better generalization ability in downstream tasks. Due to the large scale of parameters for the PTM, PTM-based CIL approaches usually adopt parameter-efficient fine-tuning~(PEFT) methods~\cite{MuNG} to learn a new task. For instance, L2P~\cite{l2p} adopt visual prompt tuning~\cite{vpt} to adapt subsequent tasks. Specifically, it constructs a prompt pool and selects a prompt via the key-query matching selection mechanism, which serves as a learnable token inserted in the embedding vector. DualPrompt~\cite{dualprompt} develops L2P by decomposing prompt into task-invariant and task-specific parts. CODA-Prompt~\cite{codaprompt} designs an attention-based weighting method to combine multiple prompts. FeCAM~\cite{fecam} proposes a feature covariance-aware Mahalanobis distance without backbone updates to handle heterogeneous class distributions. APER~\cite{aper} explores various PEFT methods and shows that prototypical classifiers serve as a strong baseline. EASE~\cite{ease} enables a model to learn new task branches and concatenates feature representations of multiple task-specific backbones. CoFiMA~\cite{cofima} introduces the fisher-weighted parameter to adaptively balance stability and plasticity via task-specific parameter importance. DGR proposes a gradient reweighting method and a distribution-aware knowledge distillation to address imbalanced catastrophic forgetting in CIL~\cite{he2024gradient, rne}. BoostCL~\cite{BMVICLR2025} enhances the model stability by theoretically justifying the effectiveness of random projection, employing a multi-view boosting strategy for a strong ensemble classifier, and integrating task-adaptive prompts with a self-improvement inference process. MiN~\cite{MiN2025} learns to generate and dynamically mix beneficial noise to mitigate catastrophic forgetting and preserve generalization. TUNA~\cite{tuna} integrates task-specific and universal adapters by effectively mitigating catastrophic forgetting and improving discrimination across tasks through specialized feature extraction and shared knowledge consolidation. MoAL~\cite{gao2025knowledge} enhances pre-trained model-based class-incremental learning with momentum-based adapter interpolation for adaptability and designs a knowledge rumination mechanism to consolidate old knowledge. SAFE~\cite{zhao2024safe} introduces complementary slow and fast parameter-efficient tuning modules to achieve a balance between stability and plasticity in continual learning with pre-trained models.

\section{The proposed approach: \textsc{Miles}} \label{sec: miles}
In this section, we provide a detailed introduction to \textsc{Miles}. Firstly, the preliminary of our work is introduced in Sec.~\ref{subsec: preliminary}, including the problem setup of CIL and the baseline of metric learning. Then, the overall structure of \textsc{Miles} is provided in Sec.~\ref{subsec: overall}. In Sec.~\ref{subsec: spp}, SPP is introduced to cascade a tiny learnable module for each task to apply metric learning to CIL without modifying the pre-trained backbone. In Sec.~\ref{subsec: sre}, a low-rank sub-network is extended to learn intermediate features from the pre-trained model to strengthen the model representational capacity. In Sec.~\ref{subsec: dr}, Distance Regularization is introduced by adding a regularization term in the optimization function to guarantee a consistent average distance of each task.

\subsection{Preliminary} \label{subsec: preliminary}
\paragraph{Class incremental learning} aims to learn a unified classifier from a data stream. Assuming that the data stream ${\cal D} = \left\{ {{{\cal D}_1},{{\cal D}_2}, \cdots ,{{\cal D}_T}} \right\}$ consists of $T$ tasks containing disjoint categories, then, ${{\cal D}_t} = \left\{ {\left( {{x_i},{y_i}} \right)} \right\}_{i = 1}^{{n_t}}$ denotes the training data for the $t^{th}$ task with $n_t$ instances, ${x_i} \in {{\cal X}_t}$ denotes the training images, ${y_i} \in {{\cal Y}_t}$ denotes the corresponding label, and ${{\cal X}_t}$, ${{\cal Y}_t}$ denote image sets and label sets for task $t$, respectively. Specially, we denote the label sets of new categories and old categories as ${{\cal Y}_n} = {{\cal Y}_t}$ and ${{\cal Y}_o} = {{\cal Y}_1} \cup  \cdots  \cup {{\cal Y}_{t - 1}}$, respectively, with ${{\cal Y}_n} \cap {{\cal Y}_o} = \emptyset $. In addition, we denote the number of categories in task $t$ as $\left| {{{\cal Y}_t}} \right| = {{\rm K}_t}$ and the number of old categories as $\left| {{{\cal Y}_o}} \right| = {\rm M} = \sum\nolimits_{i = 1}^{t - 1} {{{\rm K}_i}}$. We follow the exemplar-free setting~\cite{lwf, l2p, dualprompt, codaprompt}, where no data from previous tasks is retained in subsequent tasks. Therefore, only ${{\cal D}_t}$ can be accessed in task $t$. After learning each task, the model is tested on all seen categories.

\paragraph{Metric learning} employs the distance between the sample features and the class prototypes as the classification criterion. As shown in Eq.~\eqref{eq:1}, it embeds the image $x$ into feature vectors through the frozen backbone network.
\begin{align}
    f = {\cal F}\left( x \right) \label{eq:1}
\end{align}
Then, for the ${{\rm N}_{\rm c}}$ training samples of category $c$, the feature set $\left\{ {f_1^{\rm c}, \cdots ,f_{{{\rm N}_{\rm c}}}^{\rm c}} \right\}$ is obtained by Eq.~\eqref{eq:1}, and the prototype is calculated by
\begin{align}
    {p_{\rm c}} = \frac{1}{{{{\rm N}_{\rm c}}}}\sum\nolimits_{i = 1}^{{{\rm N}_{\rm c}}} {f_i^{\rm c}} \label{eq:2}
\end{align}
Let the number of all categories be $C$, then, a set of prototypes $\left\{ {{p_1}, \cdots ,{p_{\rm c}}} \right\}$ is obtained by Eq.~\eqref{eq:2}. For the embedded feature vector ${f_{\rm u}}$ of an unknown image, we calculate the Manhattan distance between the unknown feature and each prototype as
\begin{align}
    {l_{\rm c}} = \left| {{f_{\rm u}} - {p_{\rm c}}} \right| \label{eq:3}
\end{align}
and the category yielding to the smallest distance is taken as the inferred label. It is worth noting that applying metric learning to CIL provides a simple and effective baseline, i.e., we only need to calculate the prototype of each new category using training samples, and can directly infer class labels with the updated set of prototypes.

\subsection{Overall structure} \label{subsec: overall}
The overall structure of the proposed method is shown in Fig.~\ref{fig:overall}, which consists of a frozen pre-trained backbone network, SPP adapters, and SRE sub-networks. Given that most pre-trained models adopt the transformer architecture, we employ the classical Vision Transformer~\cite{visonTransformer} as the backbone network~\cite{cofima, ease, aper}. When the $t^{th}$ task arrives, a new SPP adapter and a SRE sub-network are added in the model to adapt for the new task. 

In the training phase, only a set of SPP adapter and SRE sub-network is involved to maintain a low training cost. The SRE sub-network learns features from samples of the current task and intermediate features from the backbone network, aiming to augment the representation capacity of the frozen backbone. Then, the output subspace feature is concatenated with the original feature of the backbone network and projected by a SPP adapter to shorten the distance to the corresponding prototypes.

In the testing phase, the backbone network is utilized only once to infer a sequence of intermediate features. Subsequently, all SRE sub-networks receive both these intermediate features and the input image, and then output subspace features. Each subspace feature is concatenated with its corresponding intermediate feature from the backbone network and fed into the dedicated SPP adapter, generating subspace projection features. Finally, Manhattan distances are computed between these subspace projection features and their respective class prototypes, and the class with the minimum distance is assigned as the final prediction.

\subsection{Subspace Prototype Projection} \label{subsec: spp}

\begin{figure}[t]
    \centering
    \includegraphics[width=3.25in]{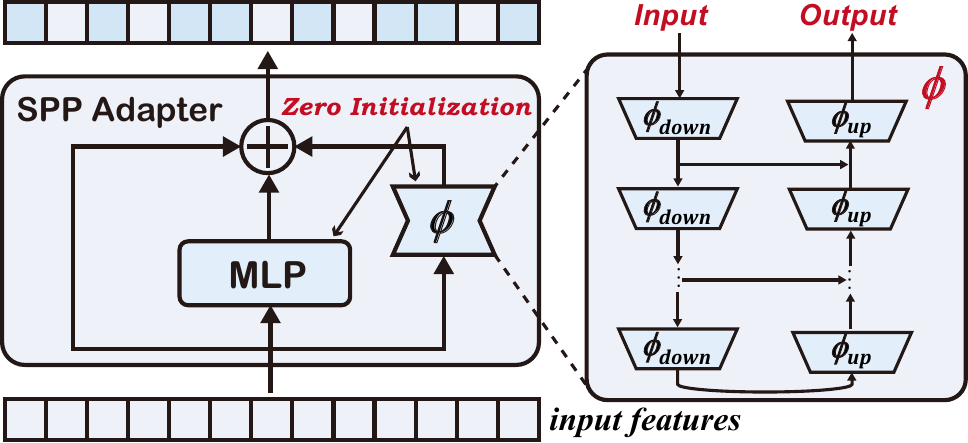}
    \caption{Structure of a SPP adapter.}
	\label{fig:spp}
\end{figure}

The motivation of integrating metric learning into CIL arises from the observation that untrained categories exhibit clustering rather than dispersion in the feature space of the pre-trained model. Therefore, as demonstrated by the baseline method in Sec.~\ref{subsec: preliminary}, metric learning can achieve CIL easily by computing prototypes of new categories and classifying the instances via distance metrics. However, such approach relies entirely on pre-trained knowledge and lacks ability to learn downstream tasks. Consequently, numerous ambiguous samples cannot be distinguished at the overlapping boundaries of different categories, limiting the model performance. To address this limitation, we aim to leverage the prior information extracted by the pre-trained model to guide the optimization of the downstream task while maintaining the scalability of the baseline.

The primary factor limiting the baseline approach is the overlapping class boundaries in the feature space. Consequently, reducing this overlap is essential for identifying ambiguous samples. Since class prototypes provide prior information for classification, we propose to shorten the distance between the sample features with the corresponding prototypes to reduce the overlapping region of different categories in the feature space. To this end, we design a lightweight learnable module, i.e., SPP adapter ${\cal A}$, cascaded after the backbone network to project the output features of the pre-trained backbone into a subspace and learn classification patterns for categories within the task. As illustrated in Fig.~\ref{fig:spp}, the SPP adapter consists of a MLP layer and a progressive down/up projection branch $\phi$. Unlike a conventional U-Net, this branch does not model spatial alignment or multi-scale spatial correspondence. Instead, it operates on the channel dimension of a 1D feature vector. The downsampling path gradually compresses the original feature into a low-dimensional bottleneck, and the upsampling path restores it to the original dimension. In this way, the adapter performs lightweight task-specific feature re-parameterization while remaining compatible with the frozen backbone representation. Its output is depicted by
\begin{align}
    {\cal A}\left( f \right) &= f + {f_{\rm mlp}} + {f_\phi } \label{eq:4}\\
    {f_{\rm mlp}} &= {W_{\rm mlp}} \cdot f + {b_{\rm mlp}} \label{eq:5}\\
    {f_\phi } &= \phi \left( f \right) \label{eq:6}
\end{align}
\noindent
where ${W_{\rm mlp}} \in {\mathbb{R}^{{\rm d_1} \times {\rm d_1}}}$, ${b_{mlp}} \in {\mathbb{R}^{{\rm d_1}}}$, ${\rm d_1}$ is the dimension of the original feature output of the frozen backbone network, and $\phi $ is composed of a down-sampling block ${\phi _{\rm down}}$ and an up-sampling block ${\phi _{\rm up}}$, i.e.,
\begin{align}
    \phi _{\rm down}^i\left( {f_{\rm in}^i} \right) &= o\left( {W_{\rm down}^if_{\rm in}^i + b_{\rm down}^i} \right) \label{eq:7}\\
    \phi _{\rm up}^i\left( {f_{\rm in}^i} \right) &= o\left( {{f_{\rm down}} + W_{\rm up}^if_{\rm in}^i + b_{\rm up}^i} \right) \label{eq:8}
\end{align}
\noindent
In the above formulae, $o\left(  \cdot  \right)$ denotes the activation function, and ${f_{\rm in}} \in {\mathbb{R}^{{\rm d_{in}}}}$ denotes the input vector. For Eq.~\eqref{eq:7}, ${W_{\rm down}} \in {\mathbb{R}^{{\rm d_{in}} \times \frac{{{\rm d_{in}}}}{r}}}$ and ${b_{\rm down}} \in {\mathbb{R}^{\frac{{{\rm d_{in}}}}{r}}}$, $r$ is a hyperparameter that determines the downsampling ratio to achieve a trade-off between performance and efficiency. For Eq.~\eqref{eq:8}, ${W_{\rm up}} \in {\mathbb{R}^{\frac{{\rm {d_{in}}}}{r} \times {\rm d_{in}}}}$ and ${b_{\rm up}} \in {\mathbb{R}^{{\rm d_{in}}}}$, ${f_{\rm down}}$ is the vector after downsampling.

The SPP adapter can be understood as a three-part residual design. First, the MLP branch performs full-dimensional linear correction on the backbone feature. Second, the down/up projection branch progressively compresses the channel dimension into a low-dimensional bottleneck and then restores it to the original dimension, so as to learn a lightweight task-specific transformation. Third, the outputs of these two learnable branches are added to the identity mapping, which keeps the adapter close to an identity transformation at initialization and enables stable adaptation during training. For clarity, we provide a concrete dimensional example of the down/up projection branch. When the backbone feature dimension is $768$ and the progressive bottleneck is configured as $\{768, 192, 48, 12\}$, the transformation path of the branch can be written as
\[
768 \rightarrow 192 \rightarrow 48 \rightarrow 12 \rightarrow 48 \rightarrow 192 \rightarrow 768.
\]
This example shows that the proposed branch performs progressive compression and restoration along the channel dimension of a 1D feature vector.

In the CIL scenario, a SPP adapter ${{\cal A}_t}$ is added to adapt to the new task when the t-th task arrives. To maintain the stability of prototypes, all weights and biases are initialized to zeros to ensure ${{\cal A}_t}\left( f \right) = f$ at the beginning of training. Then, the distance between the intra-task categories and their corresponding prototypes is optimized according to Eq.~\eqref{eq:9}.
\begin{align}
    {{\cal L}_{\rm center}}\left( {{x_i},{y_i}} \right) = \left| {{{\cal A}_t}\left( {{\cal F}\left( {{x_i}} \right)} \right) - {p_{{y_i}}}} \right| \label{eq:9}
\end{align}

\begin{figure}[t]
    \centering
    \includegraphics[width=3in]{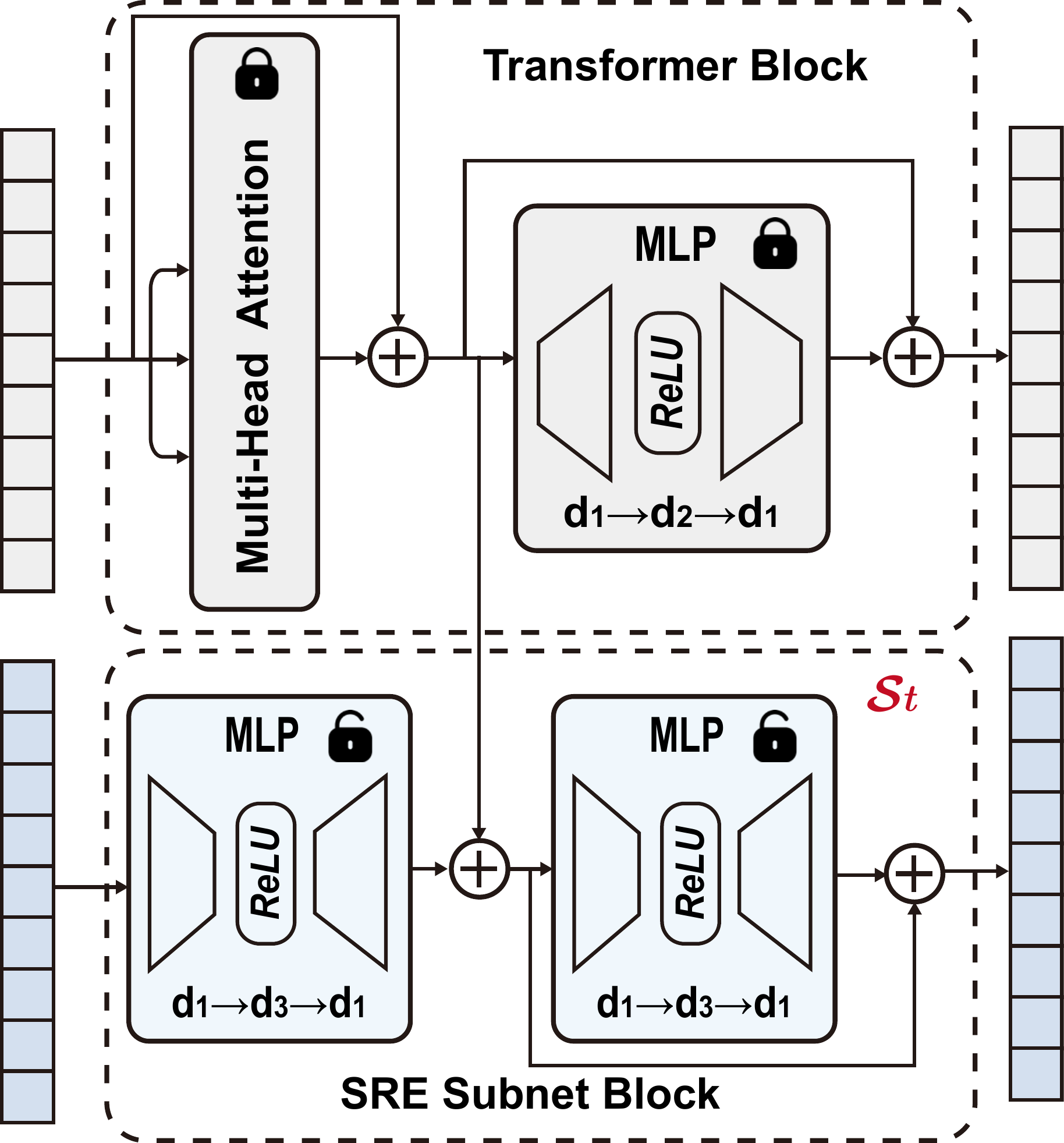}
    \caption{Structure of a SRE sub-network.}
	\label{fig:sre}
\end{figure}

Considering that merely optimizing the SPP adapter ${{\cal A}_t}$ by Eq.~\eqref{eq:9} may induce overfitting to new tasks (as all input vectors will be pulled closer to the prototypes of new categories), we propose to introduce the prior information of old categories to enhance distinguishability between the new task and old tasks. Specifically, features of old categories could be modeled by pseudo Gaussian distribution ${G_{\rm c}}\left( {{p_{\rm c}},{\sigma _{\rm c}}} \right)$, where ${p_{\rm c}} \in \left\{ {{p_1}, \cdots ,{p_{\rm M}}} \right\}$ denotes the prototype and ${\sigma _{\rm c}}$ denotes the variance of the ${\rm c}^{th}$ category. Then, pseudo features ${\hat f_{\rm c}}$ of the ${\rm c}^{th}$ category can be sampled from ${G_{\rm c}}\left( {{p_{\rm c}},{\sigma _{\rm c}}} \right)$. Then, ${{\cal L}_{\rm stable}}$ is calculated by Eq.~\eqref{eq:10} to increase the distance between the old categories and the new task prototypes in the new task feature space:
\begin{align}
    {{\cal L}_{\rm stable}}\left( {{{\hat f}_{\rm c}}} \right) = \frac{1}{{{{\rm K}_t}}}\sum\limits_{k = 1}^{{\rm{K}_t}} {\frac{1}{{\left| {{{\cal A}_t}\left( {{{\hat f}_{\rm c}}} \right) - {p_{{\rm M} + k}}} \right|}}}  \label{eq:10}
\end{align}

Finally, the training loss of the SPP strategy is defined by Eq.~\eqref{eq:11} with a hyperparameter $\beta$ for the trade-off between model stability and plasticity:
\begin{align}
    {{\cal L}_{\rm center}} + \beta  \cdot {{\cal L}_{\rm stable}} \label{eq:11}
\end{align}

The center loss minimizes the distance between intra-task features and their corresponding prototypes, while the stability loss maximizes the distance between out-of-task features and intra-task categories. By this means, we ensure that the distance from the output features to the prototype of the ground truth is less than that to other prototypes, thereby mitigating the issue of overlapping decision boundaries among different categories. In the test phase, features of the input images are extracted by the pre-trained backbone network and processed by a set of SPP adapters $\left\{ {{{\cal A}_1}, \cdots ,{{\cal A}_t}} \right\}$. Then, distances in the task subspace are calculated by Eq.~\eqref{eq:12}:
\begin{align}
    {l_{\rm c}} = \left| {{{\cal A}_i}\left( {{\cal F}\left( x \right)} \right) - {p_{\rm c}}} \right|,{\rm c} \in {{\cal Y}_i} \label{eq:12}
\end{align}
Finally, a set of distances $\left\{ {{l_1}, \cdots ,{l_{{\rm M} + {{\rm K}_t}}}} \right\}$ is obtained and the category with the smallest distance is identified as the predicted class.

\subsection{Subspace Representation Extension} \label{subsec: sre}
The SPP adapter projects original features into a task-specific subspace. Leveraging class prototype priors, it guides intra-task category feature aggregation to minimize the overlaps of decision boundaries across tasks. However, as the backbone network is entirely frozen, such learning strategy is heavily dependent on the representational capacity acquired during pre-training. Consequently, purely adopting the cascaded SPP adapter is limited to learning high-dimensional semantics from the output features of the pre-trained model, which struggles to extract additional features for new tasks directly from the original image data.

In view of this, we introduce a sub-network, i.e., SRE, to enhance the representation capacity of a new task without inducing large computational load. Fig.~\ref{fig:sre} illustrates the detailed structure of a block in the SRE sub-network. Since each block in the backbone network comprises an attention module and a feed-forward network~(FFN) layer, we utilize two low-rank MLP modules per block in the sub-network to approximate architecture of the backbone block with minimal computational overhead. The first low-rank MLP module mimics the mapping relation of the attention module, while the second one approximates the FFN layer. The output of the attention module is then connected between these two low-rank MLP layers. By this means, we enable the SRE sub-network to integrate prior knowledge from the pre-trained model and efficiently extend the feature representations for new tasks at low computational cost.

\begin{algorithm}[t]
\caption{Training pipeline of \textsc{Miles}} \label{algorithm1}
\DontPrintSemicolon 
\SetAlgoLined
\KwIn{New data ${{\cal D}_t}$, pre-trained model ${{\cal F}}$, SRE sub-network ${{\cal S}_t}$ and SPP adapter ${{\cal A}_t}$.}
\KwOut{New task prototypes $\left\{ {{p_{\rm M+1}}, \cdots ,{p_{\rm M+K_{\rm t}}}} \right\}$, updated ${{\cal S}_t}$ and ${{\cal A}_t}$}
Calculate the prototypes $\left\{ {{p_{\rm M+1}}, \cdots ,{p_{\rm M+K_{\rm t}}}} \right\}$ of new task categories using the pre-trained model\\
\For{$epoch=1$ \KwTo $epochs$}{
  Forward propagation using ${{\cal F}}$, ${{\cal S}_t}$ and ${{\cal A}_t}$\;
  Calculate the loss by Eq.~\eqref{eq:26}\;
  Update ${{\cal S}_t}$ and ${{\cal A}_t}$\;
  Recalculate the prototypes $\left\{ {{p_{\rm M+1}}, \cdots ,{p_{\rm M+{K_t}}}} \right\}$\;
}
Calculate ${{\bar l}_{\rm old}^{t}}$ by Eq.~\eqref{eq:24} and~\eqref{eq:25}\\
\Return $\left\{ {{p_{\rm M+1}}, \cdots ,{p_{\rm M+K_{\rm t}}}} \right\}$, ${{\cal S}_t}$ and ${{\cal A}_t}$
\end{algorithm}

Specifically, the pre-trained model ${\cal F} = \left\{ {{B_0},{B_1}, \cdots ,{B_L}} \right\}$ consists of an embedding layer ${B_0}$ and $L$ transformer blocks $\left\{ {{B_1}, \cdots ,{B_L}} \right\}$. In the training phase, an input image $x$ is processed by an embedding layer to generate a feature vector $x_0^{\cal F}$ according to Eq.~\eqref{eq:13}, and the entire forward propagation process of the pre-trained model is described by Eq.~\eqref{eq:13} and~\eqref{eq:14}:
\begin{align}
    {\bf x}_0^{\cal F} &= {B_0}\left( x \right) \label{eq:13} \\
    {\bf x}_i^{\cal F} &= {B_i}\left( {{\bf x}_{i - 1}^{\cal F}} \right) \label{eq:14}
\end{align}
In particular, for the $t^{th}$ task, only the $t^{th}$ SRE sub-network ${{\cal S}_t} = \left\{ {s_1^t \cdots ,s_L^t} \right\}$ is involved in the forward propagation process during training, as defined in Eq.~\eqref{eq:15} and~\eqref{eq:16}:
\begin{align}
    {\bf x}_1^t &= s_1^t\left( {{\bf x}_0^{\cal F},Attn_1^{\cal F}\left( {{\bf x}_0^{\cal F}} \right)} \right) \label{eq:15}\\
    {\bf x}_i^t &= s_i^t\left( {{\bf x}_{i - 1}^t,Attn_i^{\cal F}\left( {{\bf x}_{i - 1}^{\cal F}} \right)} \right) \label{eq:16}
\end{align}
where $Atten_i^{\cal F}$ denotes the attention module of the $i^{th}$ transformer block in the backbone network, and $s{}_i^t$ denotes the $i^{th}$ SRE block in the $t^{th}$ SRE sub-network, which is calculated by
\begin{align}
    {\bf x}_i^t = {\rm MLP}_{i,2}^t\left( {{\rm MLP}_{i,1}^t\left( {{\bf x}_{i - 1}^t} \right) + Attn_i^{\cal F}\left( {{\bf x}_{i - 1}^{\cal F}} \right)} \right) \label{eq:17}
\end{align}
where ${\rm MLP}_{i,1}^t$ and ${\rm MLP}_{i,2}^t$ denote the first and second MLP modules in the SRE block, each module consisting of a downsampling layer, an activation function, and an upsampling layer. Finally, the subspace feature output by the SRE sub-network and the original feature output by the pre-trained model are concatenated and fed into the SPP adapter ${{\cal A}_t}$ to update the model for the new task. After each training epoch, the prototypes of the new task are updated to obtain a suitable center for clustering features.

\begin{algorithm}[t]
\caption{Test pipeline of \textsc{Miles}} \label{algorithm2}
\DontPrintSemicolon 
\SetAlgoLined
\KwIn{Input image $x$, pre-trained model ${{\cal F}}$, SRE sub-networks $\left\{ {{{{\cal S}_1}}, \cdots ,{{{\cal S}_t}}} \right\}$ and SPP adapters $\left\{ {{{\cal A}_1}, \cdots ,{{\cal A}_t}} \right\}$.}
\KwOut{Prediction logits $\left\{ {l_1, \cdots ,l_{{\rm M}+{\rm K_t}}} \right\}$}
\For{$i=1$ \KwTo $t$}{
  \For{$k=1$ \KwTo ${\rm K}_i$}{
    Calculate the distance for category $c$ by: ${l_{\rm c}} = \left| {{{\cal A}_i}\left( {{\cal F}\left( x \right) \oplus {{\cal S}_i}\left( x \right)} \right) - {p_{\rm c}}} \right|,{\rm c} \in {{\cal Y}_i}$\;
  }
}
\Return Prediction logits $\left\{ {l_1, \cdots ,l_{{\rm M}+{\rm K_t}}} \right\}$
\end{algorithm}

\subsection{Distance Regularization} \label{subsec: dr}
Although the application of SPP and SRE strategies establishes a subspace for each new task to minimize task confusion, the absence of a unified distance constraint across task subspaces hinders a consistent metric criterion. This deficiency can result in a model exhibiting substantially reduced distances from projected features to prototypes in specific task subspaces relative to others, leading to a bias toward those tasks. To mitigate this limitation, we introduce two distance regularization terms to constrain the training of new tasks and maintain inter-task balance. 

The first term is the task average distance constraint, designed to ensure maximal consistency in the average distance from feature vectors to irrelevant tasks. Specifically, for a sample $\left( {{x_i},{y_i}} \right) \in {{\cal D}_t}$ from task $t$, we randomly select a previous task and project it into the subspace of task $t$, yielding the projected feature vector in Eq.~\eqref{eq:18}:
\begin{align}
    f_i^m = {{\cal A}_m}\left( {{\cal F}\left( {{x_i}} \right) \oplus {{\cal S}_m}\left( {{x_i}} \right)} \right) \label{eq:18}
\end{align}
where $m$ is the index of a random task, ${{\cal S}_m}\left( {{x_i}} \right)$ is the feature vector output by the SRE block, and $f_i^m$ is the projected feature vector in the feature subspace of task $m$. Subsequently, we compute the average distance between $f_i^m$ and the prototypes for the task $m$ by:
\begin{align}
    {\bar l_m} = \frac{1}{{{{\rm K}_m}}}\sum\nolimits_{k = 1}^{{{\rm K}_m}} {\left| {f_i^m - {p_k}} \right|} \label{eq:19}
\end{align}

Then, we sample a set of pseudo-features $\left\{ {\hat f_k^m} \right\}_{k = 1}^{{{\rm K}_m}}$ from the pseudo-Gaussian distribution ${G_k}\left( {{p_k},{\sigma _k}} \right),k \in {{\cal Y}_m}$ of each category in task $m$, and then calculate the average distance to the prototype ${p_{{y_i}}}$ in the feature subspace of task $t$ by:
\begin{align}
    {\bar l_t} = \frac{1}{{{{\rm K}_m}}}\sum\nolimits_{k = 1}^{{{\rm K}_m}} {\left| {{{\cal A}_t}\left( {\hat f_k^m} \right) - {p_{{y_i}}}} \right|} \label{eq:20}
\end{align}
Finally, the task average distance constraint is presented in Eq.~\eqref{eq:21}.
\begin{align}
    {{\cal L}_{\rm task}} = \left| {{{\bar l}_m} - {{\bar l}_t}} \right| \label{eq:21}
\end{align}
The second term is the class average distance constraint, designed to ensure consistent average distances between intra-task features and their corresponding prototypes across all task subspaces. Specifically, after the model is trained on the initial task, the average distance between samples of each class and their corresponding prototypes is first calculated as:
\begin{align}
    \bar l_{\rm old}^1 = \frac{1}{{{n_1}}}\sum\nolimits_{i = 1}^{{n_1}} {\left| {{{\cal A}_1}\left( {{\cal F}\left( {{x_i}} \right) \oplus {\cal S}\left( {{x_i}} \right)} \right) - {p_{{y_i}}}} \right|} \label{eq:22}
\end{align}
For each subsequent task $t$ ($t>1$), we compute the average distance between the features and their corresponding prototypes within each batch. This distance is then optimized to align with the class average distance from old tasks by:
\begin{align}
    {{\cal L}_{\rm class}} = \left| {{{\bar l}_{\rm old}^{t-1}} - \frac{1}{{\cal B}}\sum\nolimits_{i = 1}^{\cal B} {\left| {{{\cal A}_t}\left( {{\cal F}\left( {{x_i}} \right) \oplus {\cal S}\left( {{x_i}} \right)} \right) - {p_{{y_i}}}} \right|} } \right| \label{eq:23}
\end{align}
where ${\cal B}$ denotes the batch size. To prevent the regularization terms from impairing the model plasticity during optimization, we employ the hyperparameter $\gamma$ to regulate these two terms. Finally, the total loss is expressed as:
\begin{align}
    {{\cal L}_{{\rm{tota}}l}} = {{\cal L}_{\rm center}} + \beta  \cdot {{\cal L}_{\rm stable}} + \gamma  \cdot \left( {{{\cal L}_{\rm task}} + {{\cal L}_{\rm class}}} \right) \label{eq:26}
\end{align}

After the training phase of task $t$, the class average distance needs to be updated by:
\begin{align}
    {\bar l_{\rm new}} &= \frac{1}{{{n_t}}}\sum\nolimits_{i = 1}^{{n_t}} {\left| {{{\cal A}_t}\left( {{\cal F}\left( {{x_i}} \right) \oplus {\cal S}\left( {{x_i}} \right)} \right) - {p_{{y_i}}}} \right|} \label{eq:24}\\
    \bar l_{\rm old}^t &= \bar l_{\rm old}^{t - 1} - \frac{1}{t}\left( {\bar l_{\rm old}^{t - 1} - {{\bar l}_{\rm new}}} \right) \label{eq:25}
\end{align}

Finally, the complete training and pipelines are presented in Algorithm~\ref{algorithm1} and~\ref{algorithm2}, respectively.

\section{Experiments}
In this section, extensive experiments are conducted to validate the effectiveness of the proposed method. The detailed experimental settings are provided in Sec.~\ref{subsec:settings}. Then, the proposed method is compared with other CIL methods in six widely adopted benchmark datasets, including CIFAR-100, CUB, Omnibenchmark, ImageNet-A, ImageNet-R and FOOD-101. Comparison results and analysis are provided in Sec.~\ref{subsec:comparision}. Furthermore, ablation study is performed in Sec.~\ref{subsec:ablation} to verify the effectiveness of each component. In Sec.~\ref{subsec:vis}, more experiments are provided to validate the robustness with different learning orders and different combinations of hyper-parameters. Additionally, we also provide a comparison of the model parameters and computational overhead among different CIL methods. Finally, visualizations of the proposed method are provided to validate its working mechanism.

\begin{table}[b]
\caption{Detailed information on six datasets.}\label{tab:datasets}
\vspace{-10pt}
\centering‌
\begin{tabular}{lcrrc}
\toprule
\textbf{Dataset} & \textbf{Classes} & \textbf{Training set} & \textbf{Test set} & \textbf{Avg. size}\\
\midrule[0.5pt]
CIFAR-100~\cite{cifar100} & 100 & 50,000 & 10,000 & 32$\times$32\\
CUB-200~\cite{cub} & 200 & 9,430 & 2,358 & 467$\times$386\\
Omnibenchmark~\cite{omnibenchmark} & 300 & 89,697 & 5,985 & 764$\times$581\\
ImageNet-A~\cite{imageneta} & 200 & 5,960 & 1,515 & 443$\times$427\\
ImageNet-R~\cite{imagenetr} & 200 & 24,000 & 6,000 & 443$\times$427\\
FOOD-101~\cite{food101} & 101 & 75,750 & 25,250 & 475$\times$479\\
\bottomrule
\end{tabular}
\end{table}

\subsection{Experimental settings} \label{subsec:settings}
\paragraph{Datasets} The following experiments are conducted on six widely adopted benchmark datasets, with detailed information shown in Tab.~\ref{tab:datasets}. It is worth noting that all the six datasets are not included in the pre-training data, thus naturally avoiding data leakage. Additionally, they exhibit significant differences in image style, resolution, and object categories from the pre-training data, thereby being capable of evaluating the model continual learning capability. 

\begin{table*}[t]
\caption{The results on Cifar, Cub and Omnibenchmark datasets}\label{tab:cmp1}
\vspace{-10pt}
\centering
‌\small‌
\begin{tabular*}{\textwidth}{@{\extracolsep{\fill}}lccccccccccccc}
\toprule
\multirow{4}{*}{{\bf Methods}} & \multirow{4}{*}{{\bf Exemplars}} & \multicolumn{4}{c}{CIFAR} & \multicolumn{4}{c}{CUB} & \multicolumn{4}{c}{Omnibenchmark}\\
  \cmidrule(lr){3-14}
  & & \multicolumn{2}{c}{\textit{T}=10} & \multicolumn{2}{c}{\textit{T}=50} & \multicolumn{2}{c}{\textit{T}=10} & \multicolumn{2}{c}{\textit{T}=50} & \multicolumn{2}{c}{\textit{T}=10} & \multicolumn{2}{c}{\textit{T}=50}\\
  \cmidrule(lr){3-4} \cmidrule(lr){5-6} \cmidrule(lr){7-8} \cmidrule(lr){9-10} \cmidrule(lr){11-12} \cmidrule(lr){13-14}
 & & $\bar A\uparrow$ & $F\downarrow$ & $\bar A\uparrow$ & $F\downarrow$ & $\bar A\uparrow$ & $F\downarrow$ & $\bar A\uparrow$ & $F\downarrow$ & $\bar A\uparrow$ & $F\downarrow$ & $\bar A\uparrow$ & $F\downarrow$\\
 \midrule
Joint$^\dag$ & \ding{56} & \multicolumn{2}{c}{92.94} & \multicolumn{2}{c}{92.94} & \multicolumn{2}{c}{93.53} & \multicolumn{2}{c}{93.53} & \multicolumn{2}{c}{82.59} & \multicolumn{2}{c}{82.59}\\
Finetune & \ding{56} & 79.24 & 22.58 & 45.03 & 63.72 & 67.32 & 41.32 & 38.13 & 74.15 & 66.86 & 32.03 & 47.28 & 56.93\\
Replay & \ding{52} & 85.06 & 15.16 & 82.00 & 24.90 & 87.82 & 12.57 & 84.55 & 20.75 & 76.05 & 14.88 & 75.21 & 18.54\\
\midrule[0.5pt]
iCaRL~\cite{icarl} & \ding{52} & 87.34 & 13.69 & 84.97 & 20.27 & 88.96 & 11.21 & 87.66 & 17.58 & 77.04 & 13.80 & 76.53 & 17.19\\
DER~\cite{der} & \ding{52} & 88.48 & 9.30 & 85.99 & 18.63 & 88.93 & 11.24 & 87.80 & 17.30 & 78.03 & 13.02 & 77.89 & 15.38\\
L2P~\cite{l2p} & \ding{56} & 85.92 & 13.75 & 74.29 & 31.11 & 84.29 & 19.02 & 78.51 & 28.94 & 72.91 & 19.30 & 69.81 & 23.49\\
DualPrompt~\cite{dualprompt} & \ding{56} & 88.44 & 9.40 & 73.66 & 28.97 & 84.39 & 20.08 & 78.06 & 28.73 & 77.38 & 15.14 & 71.28 & 22.66\\
FOSTER~\cite{foster} & \ding{52} & 89.75 & 8.74 & 87.98 & 15.28 & 89.36 & 9.69 & 88.39 & 16.20 & 79.40 & 10.93 & 79.60 & 12.84\\
CODA-Prompt~\cite{codaprompt} & \ding{56} & 89.28 & 8.50 & 69.54 & 37.28 & 84.15 & 19.53 & 75.66 & 31.70 & 78.96 & 12.58 & 70.50 & 22.09\\
FeCAM~\cite{fecam} & \ding{56} & 90.83 & 6.74 & 88.43 & 10.06 & 91.23 & 7.15 & 91.16 & 7.65 & 82.46 & 8.35 & 80.03 & 11.36\\
BEEF~\cite{beef} & \ding{52} & 85.10 & 7.84 & -- & -- & 90.00 & 9.06 & 88.89 & 15.38 & 80.22 & 9.44 & 80.11 & 11.98\\
APER~\cite{aper} & \ding{56} & 90.98 & 6.35 & 87.97 & 11.03 & 88.32 & 7.78 & 88.42 & 8.24 & 80.05 & 9.29 & 80.27 & 9.59\\
EASE~\cite{ease} & \ding{56} & 92.11 & 5.22 & 84.31 & 18.47 & 90.12 & 9.77 & 88.28 & 7.06 & 75.93 & 14.00 & 73.56 & 17.07\\
CoFiMA~\cite{cofima} & \ding{56} & 92.51 & 4.87 & 82.64 & 19.70 & 87.22 & 13.93 & 82.72 & 21.53 & 81.69 & 9.21 & 72.11 & 19.84\\
MTD~\cite{mtd} & \ding{52} & 91.72 & 6.35 & -- & -- & 90.38 & 8.49 & 88.48 & 10.06 & 80.62 & 8.54 & 81.59 & 9.89\\
DGR~\cite{he2024gradient} & \ding{52} & \underline{93.05} & 4.85 & 89.74 & 9.74 & 91.15 & 7.61 & 91.55 & 6.93 & \underline{83.14} & \underline{6.46} & \underline{82.70} & 
\underline{8.30}\\
SAFE~\cite{zhao2024safe} & \ding{56} & 92.56 & \underline{3.38} & 88.47 & 8.10 & 90.32 & \underline{4.73} & 89.30 & 5.57 & 77.21 & 13.29 & 74.39 & 18.15\\
TUNA~\cite{tuna} & \ding{56} & 91.10 & 3.40 & \underline{91.08} & \textbf{4.17} & \underline{92.55} & 5.29 & \underline{92.65} & \underline{5.09} & 82.91 & 13.12 & 81.36 & 11.44\\
\midrule[0.5pt]
\textsc{Miles} & \ding{56} & \textbf{93.93} & \textbf{3.12} & \textbf{91.90} & \underline{6.03} & \textbf{93.02} & \textbf{4.57} & \textbf{92.86} & \textbf{4.94} & \textbf{85.02} & \textbf{4.04} & \textbf{84.84} & \textbf{5.45}\\
\bottomrule
\multicolumn{10}{l}{$^\dag$ denotes the result of joint training for all tasks, and only the final accuracy is reported.}\\
\end{tabular*}
\end{table*}

\begin{table*}[t]
\caption{The results on Imagenet-A/R and Food datasets}\label{tab:cmp2}
\centering
‌\small
\begin{tabular*}{\textwidth}{@{\extracolsep{\fill}}lccccccccccccc}
\toprule
\multirow{4}{*}{{\bf Methods}} & \multirow{4}{*}{{\bf Exemplars}} & \multicolumn{4}{c}{ImageNet-A} & \multicolumn{4}{c}{ImageNet-R} & \multicolumn{4}{c}{FOOD}\\
  \cmidrule(lr){3-14}
  & & \multicolumn{2}{c}{\textit{T}=10} & \multicolumn{2}{c}{\textit{T}=50} & \multicolumn{2}{c}{\textit{T}=10} & \multicolumn{2}{c}{\textit{T}=50} & \multicolumn{2}{c}{\textit{T}=10} & \multicolumn{2}{c}{\textit{T}=50}\\
  \cmidrule(lr){3-4} \cmidrule(lr){5-6} \cmidrule(lr){7-8} \cmidrule(lr){9-10} \cmidrule(lr){11-12} \cmidrule(lr){13-14}
 & & $\bar A\uparrow$ & $F\downarrow$ & $\bar A\uparrow$ & $F\downarrow$ & $\bar A\uparrow$ & $F\downarrow$ & $\bar A\uparrow$ & $F\downarrow$ & $\bar A\uparrow$ & $F\downarrow$ & $\bar A\uparrow$ & $F\downarrow$\\
 \midrule
Joint$^\dag$ & \ding{56} & \multicolumn{2}{c}{54.66} & \multicolumn{2}{c}{54.66} & \multicolumn{2}{c}{84.50} & \multicolumn{2}{c}{84.50} & \multicolumn{2}{c}{92.28} & \multicolumn{2}{c}{92.28}\\
Finetune & \ding{56} & 37.45 & 37.68 & 21.69 & 46.13 & 71.46 & 27.42 & 48.12 & 50.83 & 77.07 & 29.39 & 43.15 & 61.40\\
Replay & \ding{52} & 39.28 & 30.43 & 23.11 & 42.85 & 76.42 & 18.33 & 69.42 & 30.42 & 79.17 & 22.00 & 75.06 & 26.17\\
\midrule[0.5pt]
iCaRL~\cite{icarl} & \ding{52} & 39.93 & 29.04 & 23.27 & 41.95 & 78.31 & 15.57 & 72.01 & 26.92 & 81.16 & 19.65 & 77.18 & 23.02\\
DER~\cite{der} & \ding{52} & 43.57 & 24.67 & 37.05 & 32.22 & 78.88 & 14.94 & 74.20 & 24.73 & 85.29 & 13.77 & 81.70 & 17.46\\
L2P~\cite{l2p} & \ding{56} & 54.46 & 12.33 & 41.15 & 23.59 & 67.76 & 24.72 & 58.02 & 36.18 & 83.32 & 17.35 & 59.66 & 42.02\\
DualPrompt~\cite{dualprompt} & \ding{56} & 56.18 & 10.82 & 35.93 & 29.05 & 71.39 & 18.00 & 55.54 & 37.11 & 86.65 & 12.26 & 67.29 & 33.48\\
FOSTER~\cite{foster} & \ding{52} & 46.68 & 20.56 & 39.41 & 29.30 & 79.94 & 13.79 & 77.37 & 20.08 & 86.09 & 12.31 & 82.81 & 15.93\\
CODA-Prompt~\cite{codaprompt} & \ding{56} & 57.79 & 9.77 & 35.69 & 31.09 & 75.70 & 14.18 & 55.98 & 39.07 & 89.21 & 9.14 & 62.88 & 40.71\\
FeCAM~\cite{fecam} & \ding{56} & 58.46 & 8.25 & 52.22 & 12.29 & 79.02 & 11.97 & 63.75 & 28.65 & 89.47 & 7.46 & 86.68 & 12.20\\
BEEF~\cite{beef} & \ding{52} & 47.04 & 20.01 & 40.97 & 27.11 & 80.67 & 12.71 &76.58 & 21.32 & 87.40 & 10.46 & 83.75 & 14.73\\
APER~\cite{aper} & \ding{56} & 60.26 & \underline{5.75} & 57.70 & 7.77 & 75.45 & 17.18 & 69.29 & 23.38 & 88.49 & 8.68 & 87.66 & 10.86\\
EASE~\cite{ease} & \ding{56} & 59.09 & 9.30 & 58.44 & 6.73 & 81.75 & 8.30 & 76.51 & 15.98 & 88.63 & 9.02 & 85.61 & 13.53\\
CoFiMA~\cite{cofima} & \ding{56} & 58.52 & 7.06 & 43.53 & 23.46 & 82.05 & 8.07 & 71.46 & 26.22 & 90.83 & 6.44 & 77.97 & 24.31\\
MTD~\cite{mtd} & \ding{52} & 49.18 & 17.52 & 37.88 & 31.54 & 82.81 & 9.00 & 77.75 & 19.33 & 88.35 & 9.38 & 84.58 & 13.75\\
DGR~\cite{he2024gradient} & \ding{52} & 43.66 & 27.80 & 36.33 & 33.40 & 83.85 & 7.73 & \underline{80.57} & 13.50 & \underline{91.15} & 5.81 & \underline{89.21} & \underline{8.54}\\
SAFE~\cite{zhao2024safe} & \ding{56} & 54.45 & 9.23 & 35.01 & 29.33 & 78.14 & \underline{7.41} & 63.75 & 21.57 & 89.75 & \underline{5.44} & 87.82 & 9.40\\
TUNA~\cite{tuna} & \ding{56} & \underline{61.34} & 6.71 & \underline{62.54} & \underline{5.29} & \underline{84.17} & 8.03 & 64.53 & \underline{12.35} & 85.36 & 26.27 & 79.62 & 24.60\\
\midrule[0.5pt]
\textsc{Miles} & \ding{56} & \textbf{62.33} & \textbf{2.92} & \textbf{62.67} & \textbf{2.45} & \textbf{84.90} & \textbf{5.87} & \textbf{81.90} & \textbf{9.96} & \textbf{93.23} & \textbf{3.96} & \textbf{90.75} & \textbf{5.74}\\
\bottomrule
\multicolumn{10}{l}{$^\dag$ denotes the result of joint training for all tasks, and only the final accuracy is reported.}\\
\end{tabular*}
\end{table*}

\begin{figure*}[t]
\centering
\subfloat[\scriptsize{CIFAR}\label{subfig:comparison-10steps(a)}]{\includegraphics[width=2.35in]{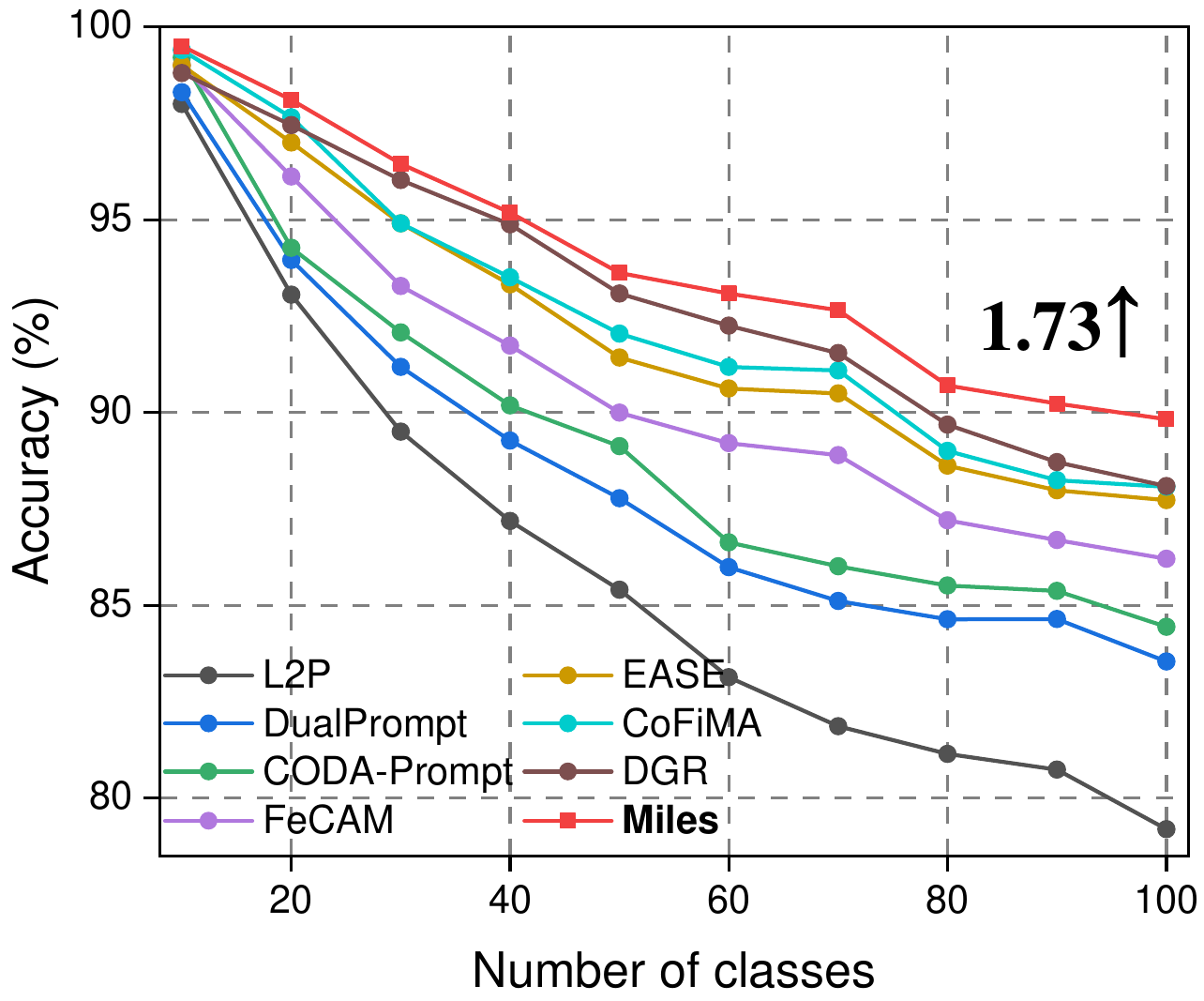}}
\hfill
\subfloat[\scriptsize{CUB}\label{subfig:comparison-10steps(b)}]{\includegraphics[width=2.35in]{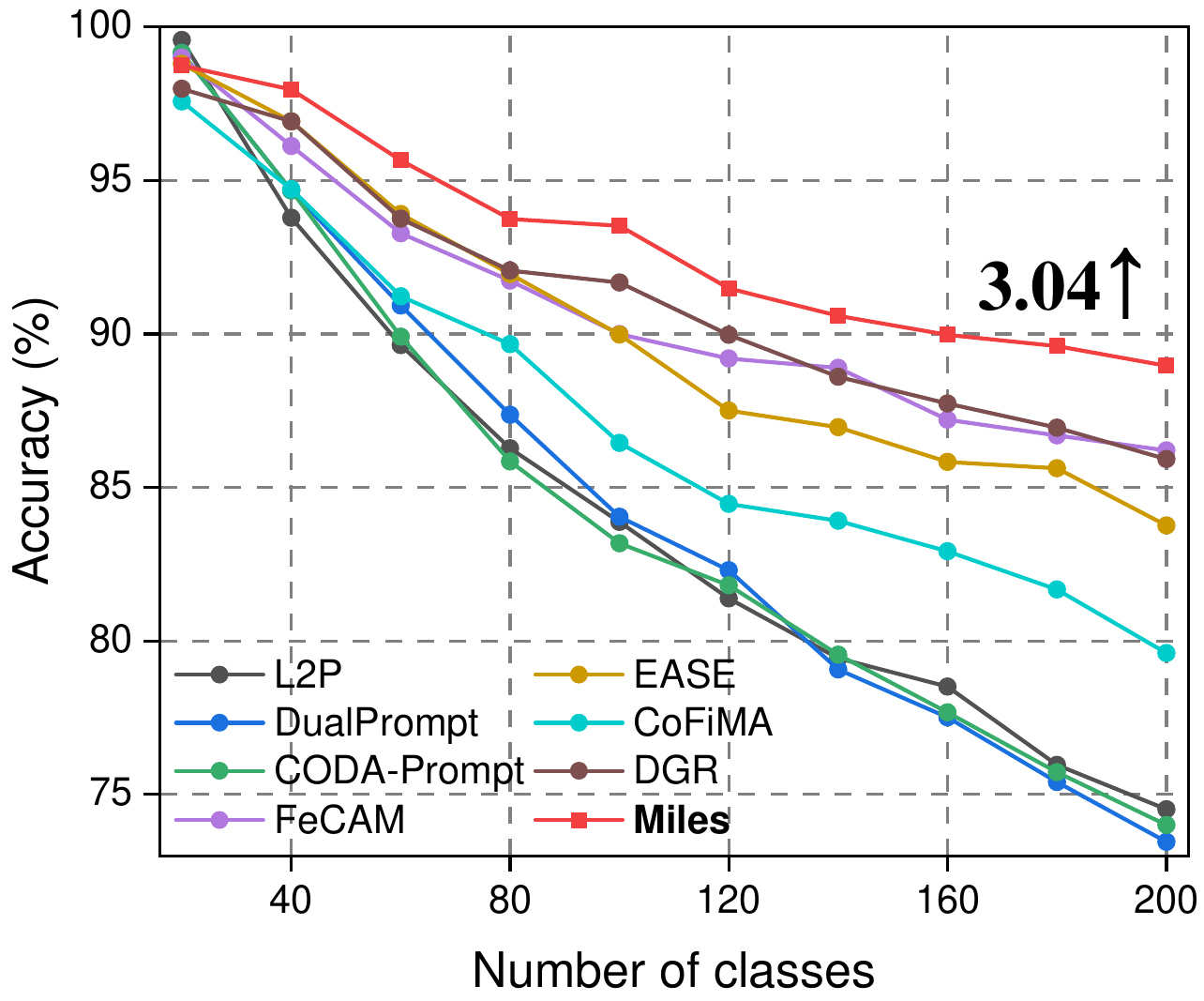}}
\hfill
\subfloat[\scriptsize{Omnibenchmark}\label{subfig:comparison-10steps(c)}]{\includegraphics[width=2.35in]{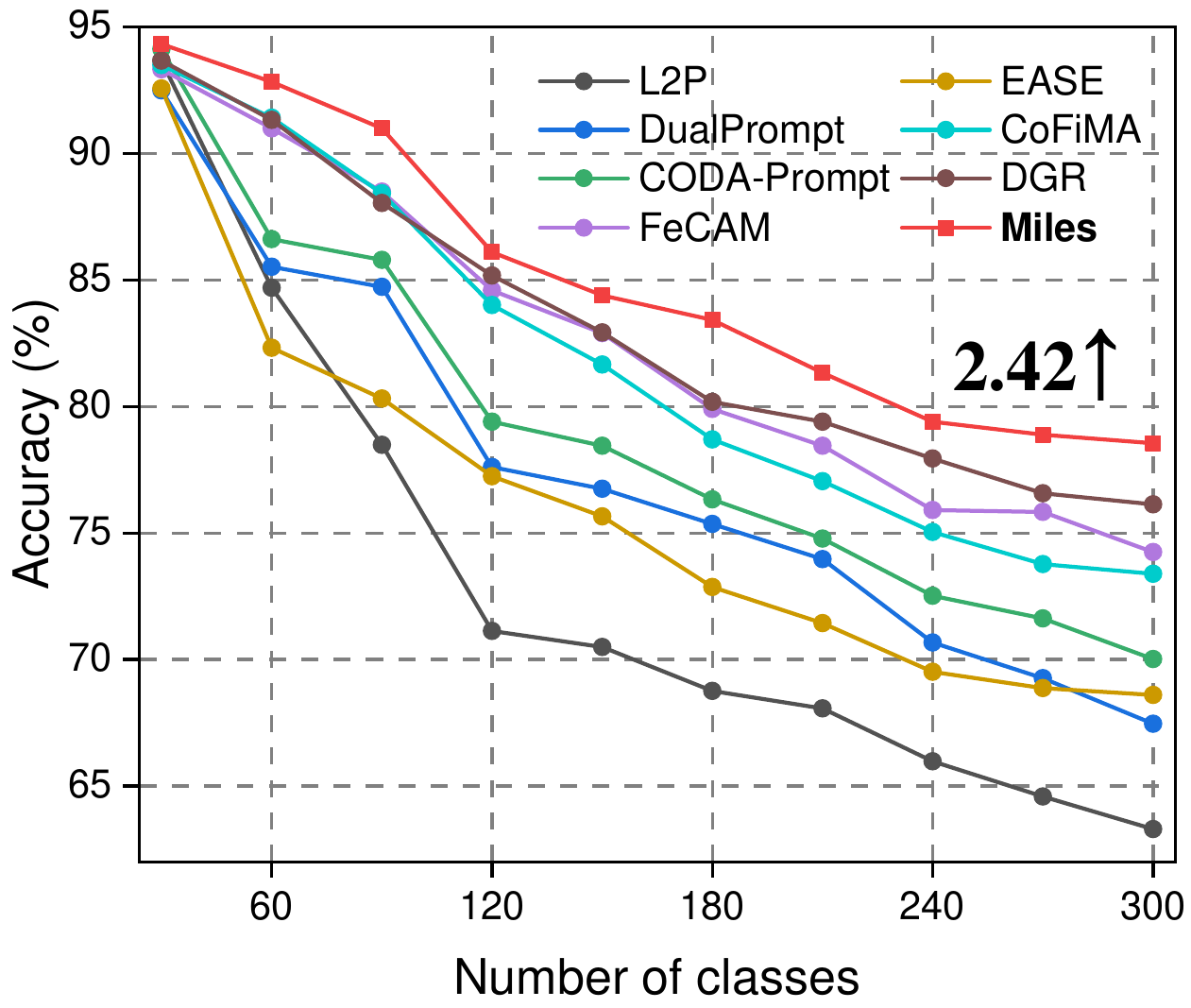}}
\\
\subfloat[\scriptsize{ImageNet-A}\label{subfig:comparison-10steps(d)}]{\includegraphics[width=2.35in]{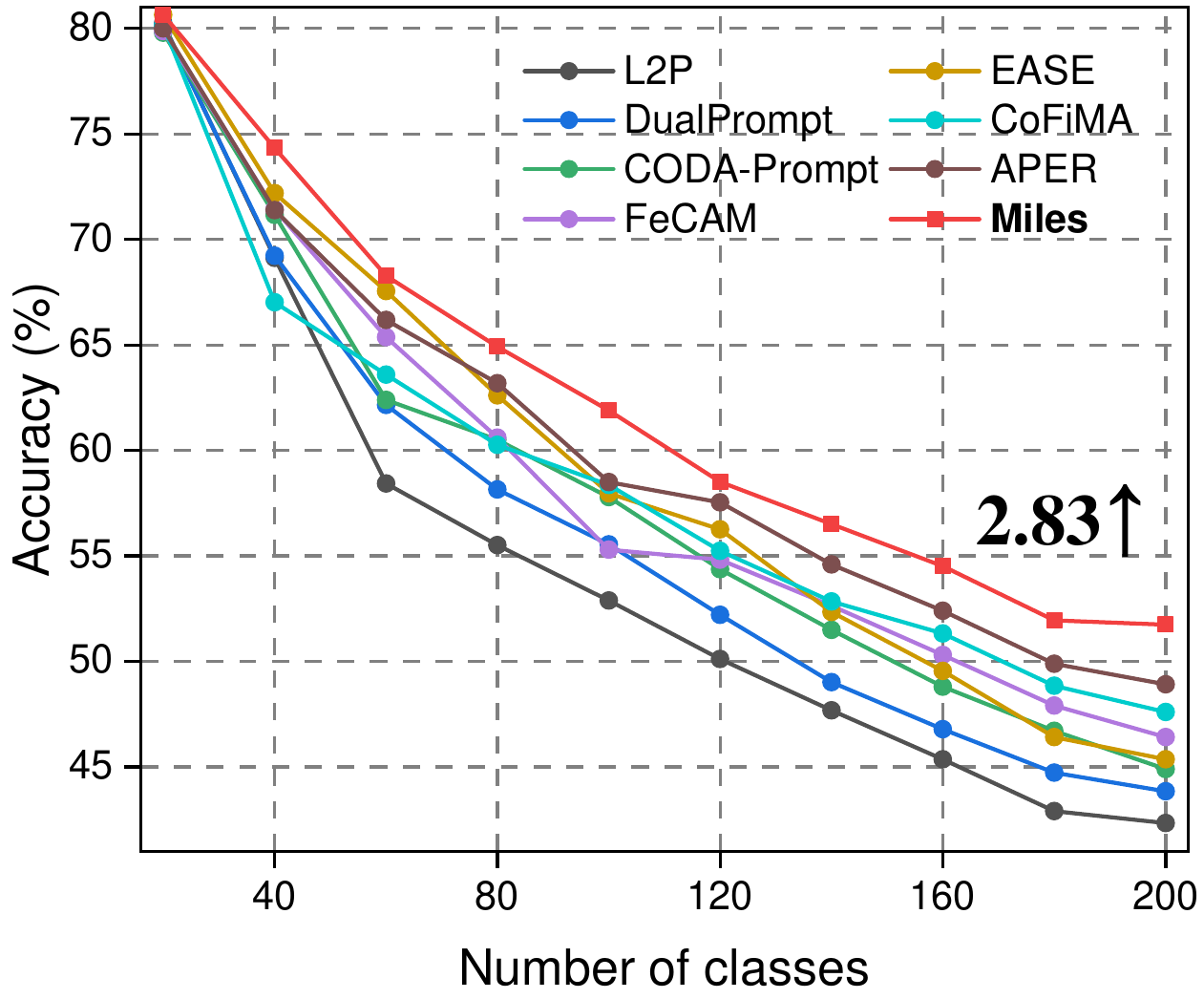}}
\hfill
\subfloat[\scriptsize{ImageNet-R}\label{subfig:comparison-10steps(e)}]{\includegraphics[width=2.35in]{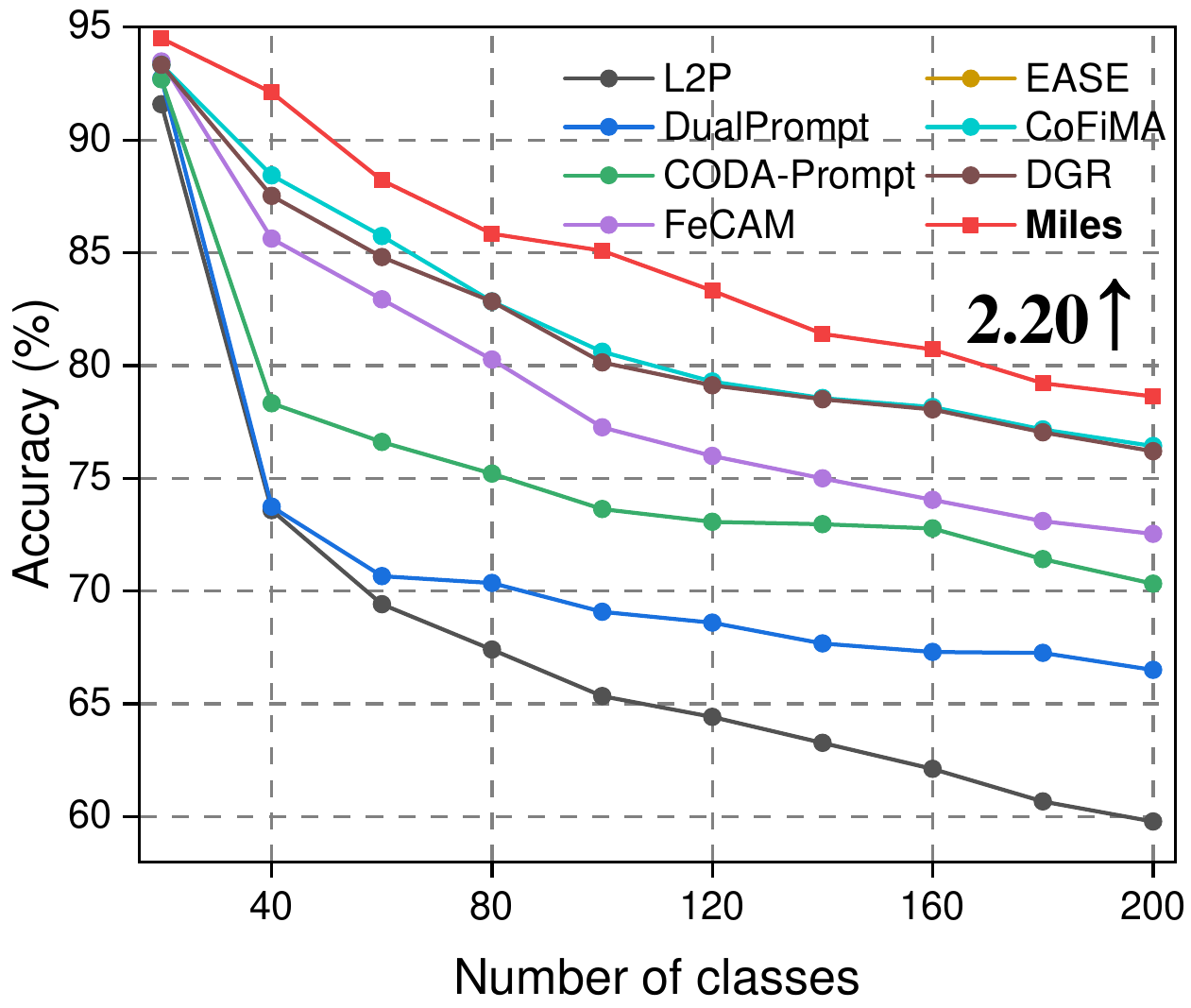}}
\hfill
\subfloat[\scriptsize{FOOD}\label{subfig:comparison-10steps(f)}]{\includegraphics[width=2.35in]{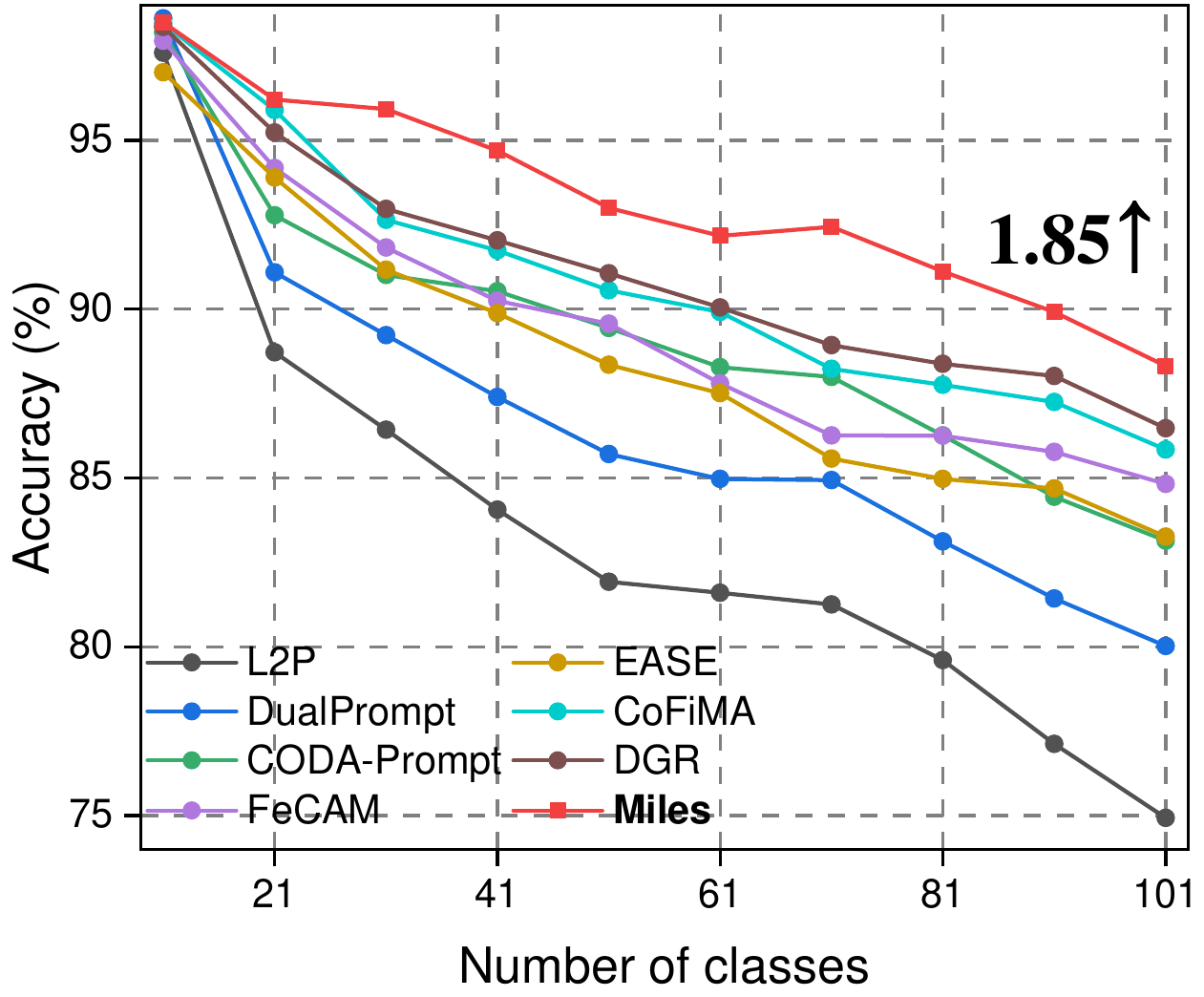}}
\caption{Incremental trends on six different datasets with the 10-step setting.}
\label{fig:comparison-10steps}
\end{figure*}

\paragraph{Protocols} Following recent work on CIL, we evenly split a dataset into $T$ tasks, denoted as $T$ steps, where the model learns an equal number of categories per step. Specifically, we set $T$ to 10 or 50 to evaluate model performance under short and long task sequences. Consistent with~\cite{icarl, l2p, codaprompt, ease}, the learning order is determined by the random seed of 1993. Each experiment is repeated three times, with results averaged for final reporting.

\begin{figure*}[t]
\centering
\subfloat[\scriptsize{CIFAR}\label{subfig:comparison-50steps(a)}]{\includegraphics[width=2.35in]{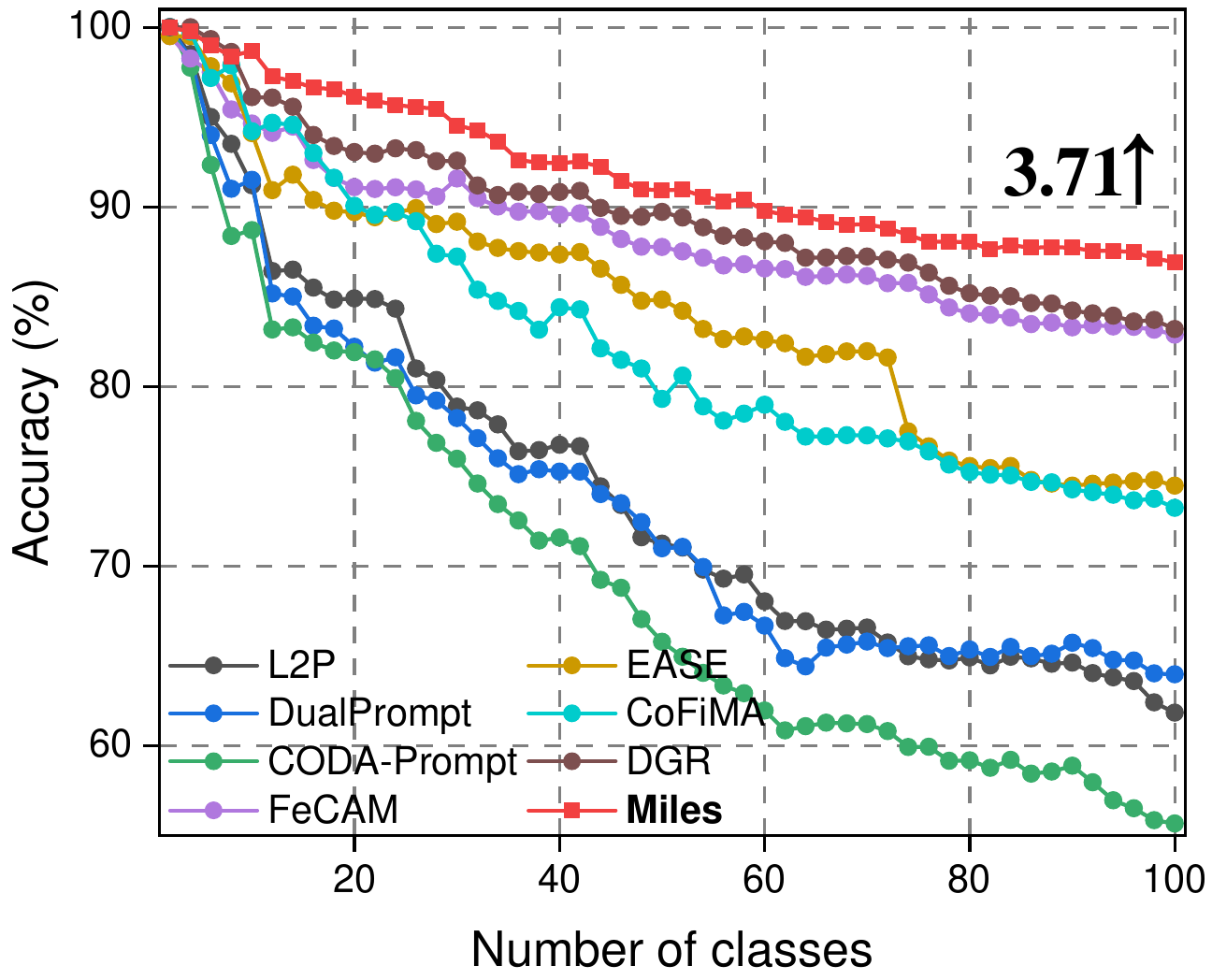}}
\hfill
\subfloat[\scriptsize{CUB}\label{subfig:comparison-50steps(b)}]{\includegraphics[width=2.35in]{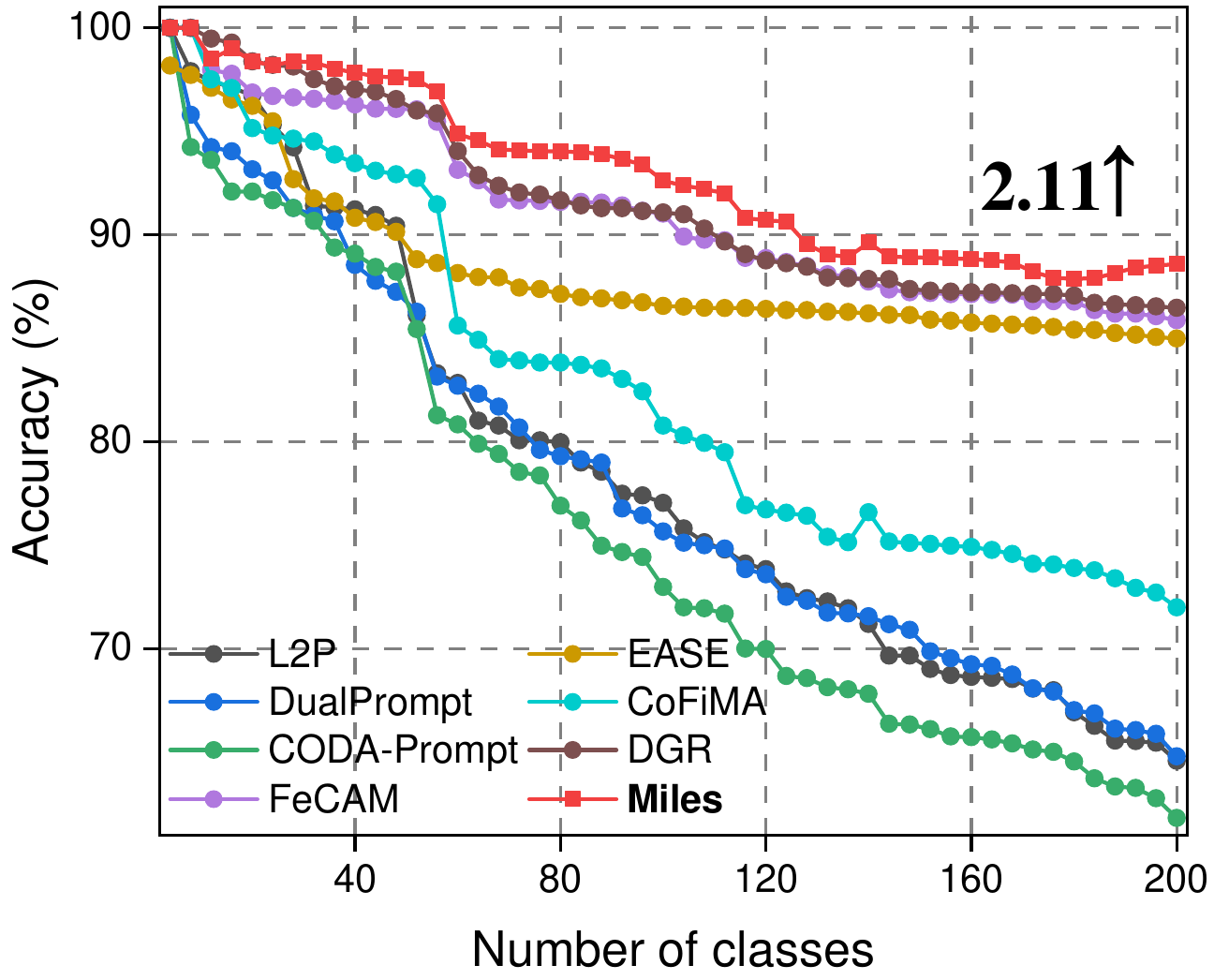}}
\hfill
\subfloat[\scriptsize{Omnibenchmark}\label{subfig:comparison-50steps(c)}]{\includegraphics[width=2.35in]{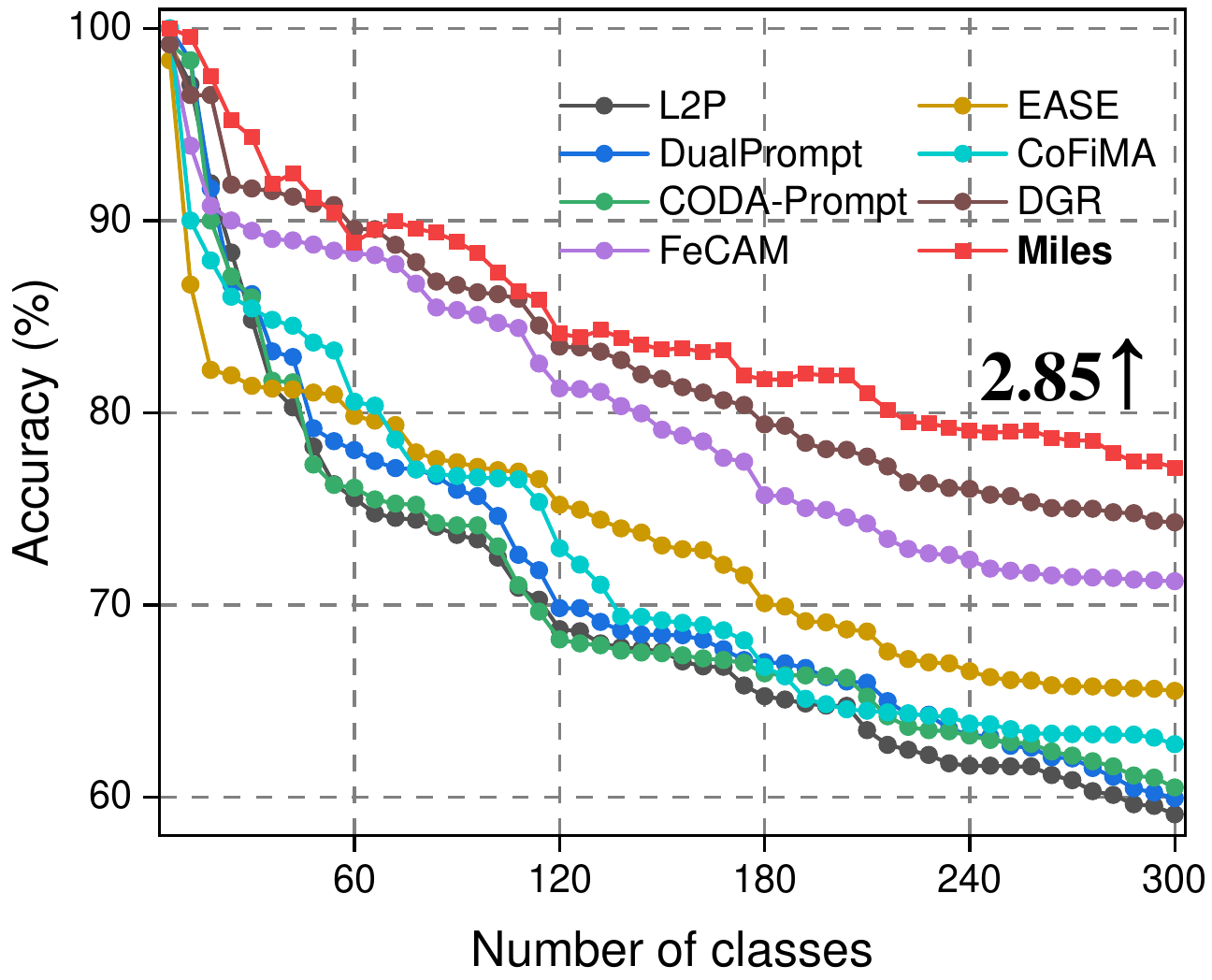}}
\\
\subfloat[\scriptsize{ImageNet-A}\label{subfig:comparison-50steps(d)}]{\includegraphics[width=2.35in]{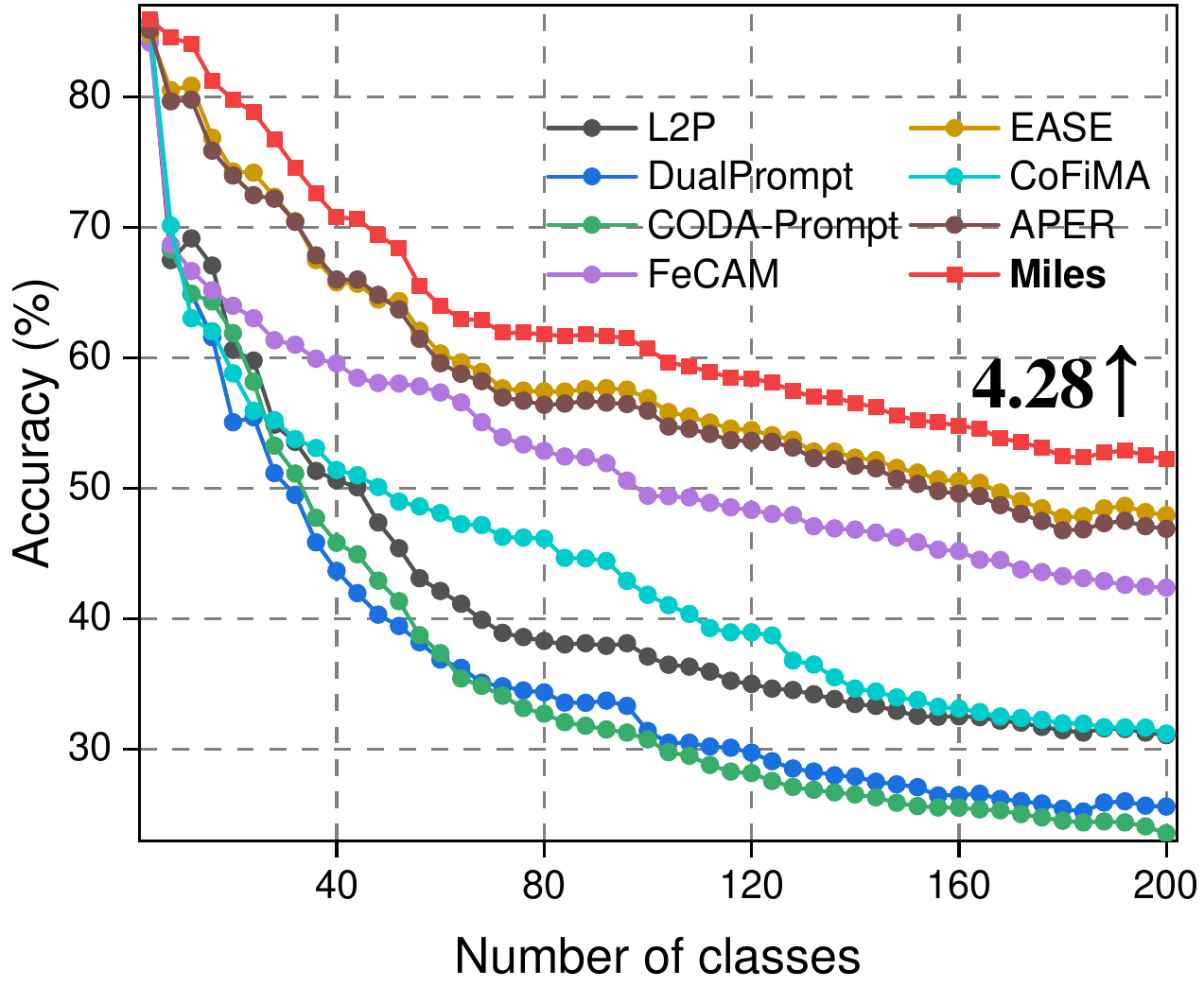}}
\hfill
\subfloat[\scriptsize{ImageNet-R}\label{subfig:comparison-50steps(e)}]{\includegraphics[width=2.35in]{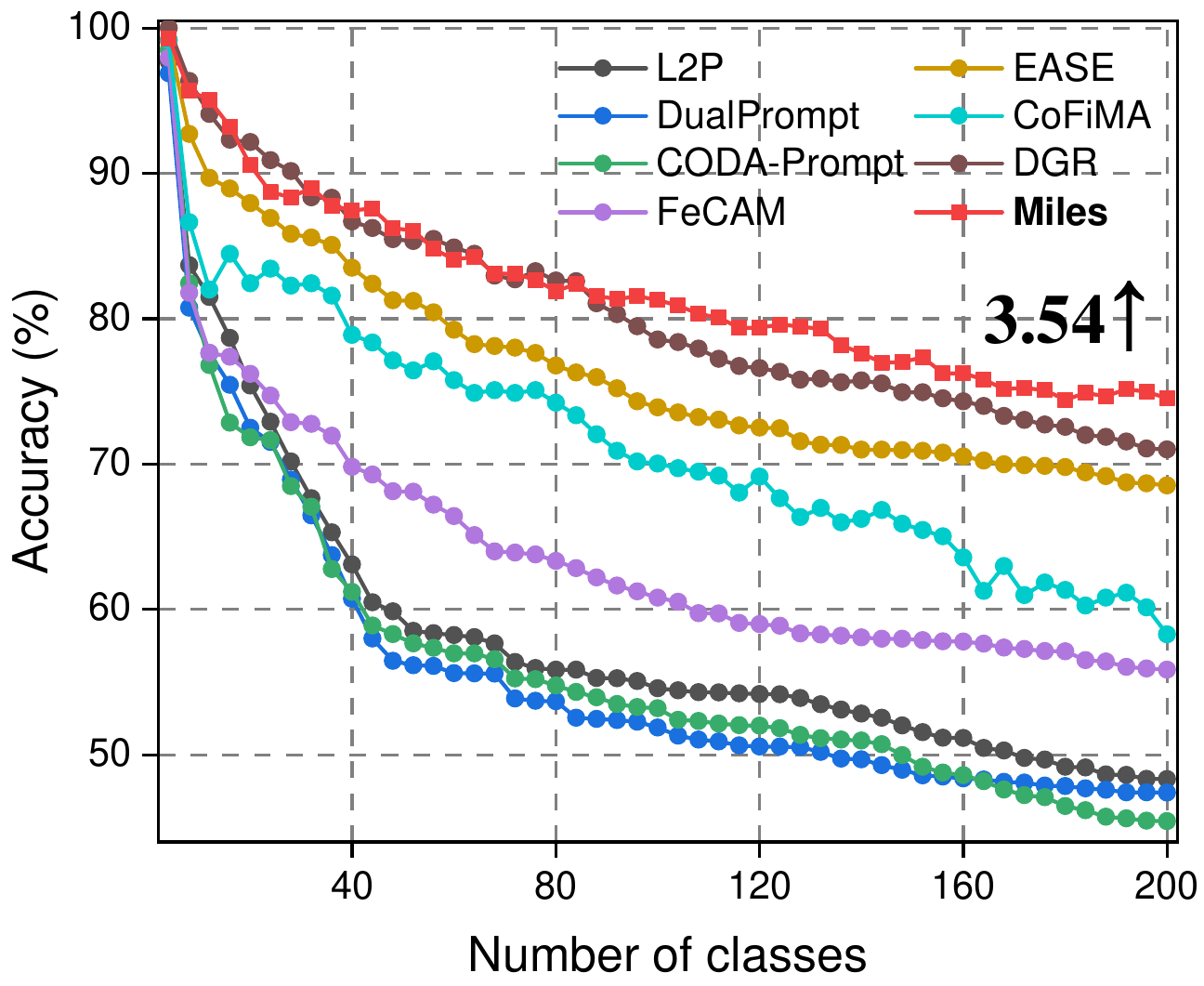}}
\hfill
\subfloat[\scriptsize{FOOD}\label{subfig:comparison-50steps(f)}]{\includegraphics[width=2.35in]{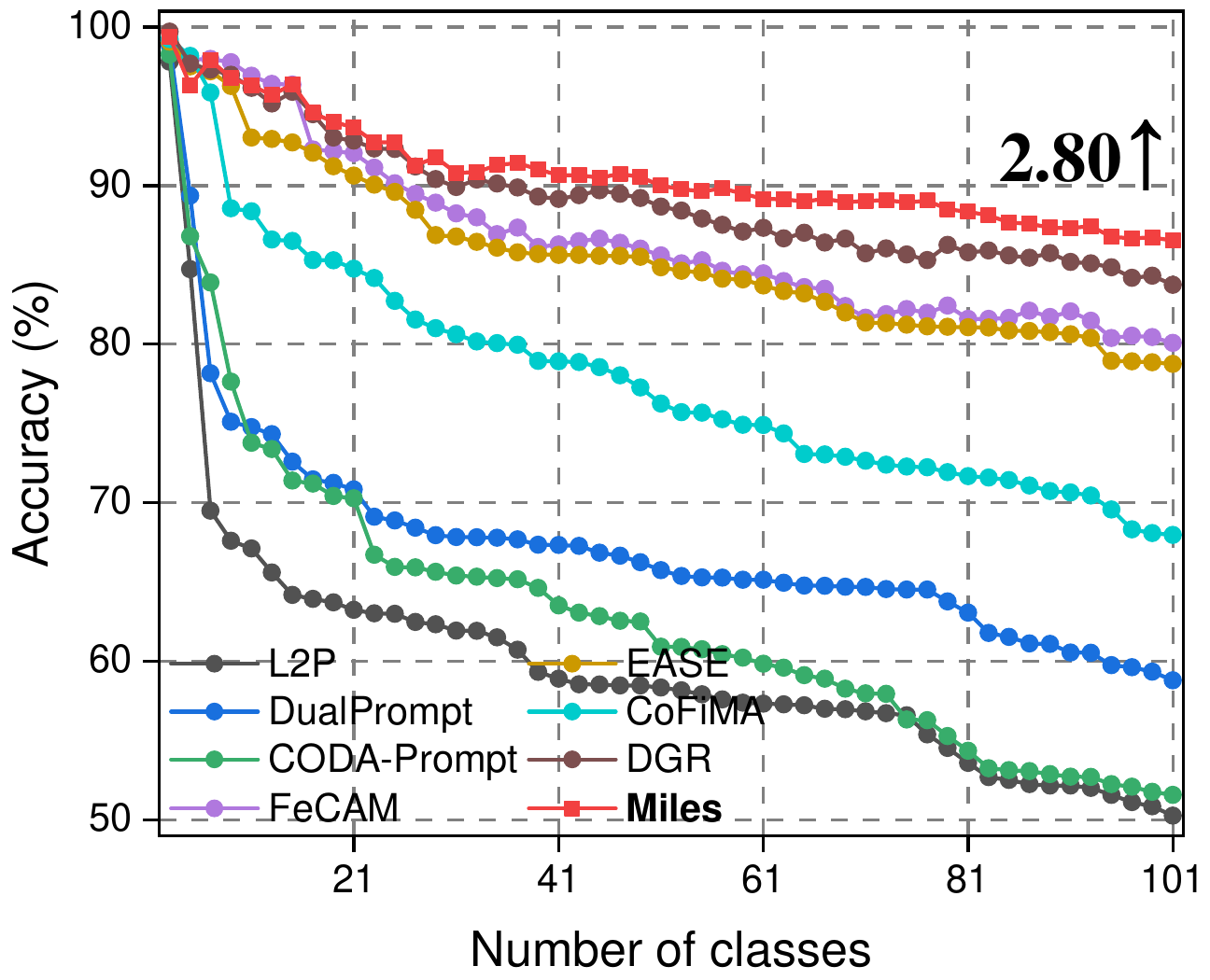}}
\caption{Incremental trends on six different datasets with the 50-step setting.}
\label{fig:comparison-50steps}
\end{figure*}

\paragraph{Evaluation metric} After learning the task $t$, we validate the model performance on all known categories. Assuming that ${A_t}$ represents the accuracy on the test set of all known categories, then, we can obtain a set of task accuracy $\left\{ {{A_1}, \cdots ,{A_T}} \right\}$ and adopt $\bar A = \frac{1}{T}{\sum\nolimits_{t = 1}^T A _t}$ for performance evaluation. Furthermore, we utilize the entire dataset to jointly train the model and subsequently compute the overall classification accuracy, denoted as ${A_{\rm joint}}$. Then, we employ the forgetting rate $F = {A_{\rm joint}} - {A_T}$ to quantify the forgetting of the CIL method compared to joint training.

\paragraph{Comparison methods} We first evaluate the performance upper bound for CIL, i.e., ${A_{\rm joint}}$, by training the model on all the available data. Subsequently, we assess two baseline methods, i.e., Finetune and Replay. Among them, Finetune learns the new task data by adjusting all parameters of the old model directly, which leads to severe catastrophic forgetting; Replay maintains a limited-capacity exemplar set to preserve a subset of old samples, and trains the model with both the exemplar set and new task data. In particular, we compare seven exemplar-free methods, i.e., L2P~\cite{l2p}, DualPrompt~\cite{dualprompt}, CODA-Prompt~\cite{codaprompt}, FeCAM~\cite{fecam}, APER~\cite{aper}, EASE~\cite{ease} and CoFiMA~\cite{cofima}; and six exemplar-based methods, i.e., iCaRL~\cite{icarl}, DER~\cite{der}, FOSTER~\cite{foster}, BEEF~\cite{beef}, MTD~\cite{mtd} and DGR~\cite{he2024gradient}, with the proposed approach. To ensure a fair comparison, all methods utilize the same pre-trained backbone network ViT-B/16-IN21K~\cite{visonTransformer}. The size of the exemplar set is fixed to 20 per category, except for ImageNet-A, which adopts 2 per category due to the limited sample number.

\paragraph{Implementation details} The backbone network ViT-B/16-IN21K is pre-trained on ImageNet-21K, and the weights are available from the timm library. To accommodate the pre-trained model, all input images are resized to 224$\times$224 pixels and then fed into the backbone network. To ensure the fairness, we reproduce the comparison methods and \textsc{Miles} with the same hyper-parameters and pretrained weight. All models are trained for 20 epochs in each task. The learning rate is initialized as 0.001 and decreased to zero with a cosine annealing scheduler. The batch size is set to 128 for the six datasets. The SGD optimizer is deployed, where the momentum factor is set to 0.9 and weight decay is set to 0.0005. In addition, the method-specific hyperparameters are kept as default without modification. For some architecture-based method such as DER, we only replace the backbone based on their original implementations and directly extend the complete pretrained network. For \textsc{Miles}, we set ${{\rm d}_3} = 16$ to maintain a low rank for each block of the SRE sub-network and set $r = 4$ to gradually adjust the dimension of the hidden vector in each SPP adapter from 768 to 12, i.e., $\left\{ {768,192,48,12} \right\}$.  The hyper-parameters $\beta$ and $\gamma$ are set to 0.1 and 0.001, respectively.

\subsection{Comparison with other approaches} \label{subsec:comparision}
To illustrate the superiority of the proposed method, we compare \textsc{Miles} with other CIL approaches on six different datasets. In this context, a higher average accuracy rate or a lower forgetting rate indicates better performance. Tab.~\ref{tab:cmp1} and~\ref{tab:cmp2} present the average accuracy and forgetting rates for each method under the settings of $T$=10 and $T$=50. For each setting, the best value is highlighted in bold, and the second-best result, i.e., the runner-up, is indicated by an underline. Furthermore, Fig.~\ref{fig:comparison-10steps} and Fig.~\ref{fig:comparison-50steps} present the incremental trends for each method under the settings of $T$=10 and $T$=50, respectively. In each trend graph, we annotate the accuracy gains of \textsc{Miles} relative to the suboptimal method after learning all tasks. Detailed analyses of the comparison experiments for each dataset are given below.

\paragraph{CIFAR} As illustrated in Tab.~\ref{tab:cmp1}, the average accuracy and forgetting rate of \textsc{Miles} in the setting of $T$=10 are 93.93\% and 3.12\%, respectively, for the CIFAR-100 dataset. Specifically, compared with the runner-up method, DGR, the average accuracy increases by 0.88\% and the forgetting rate decreases by 1.73\%. Furthermore, the average accuracy and forgetting rate of \textsc{Miles} in the setting of $T$=50 are 91.90\% and 6.03\%, respectively. \textsc{Miles} outperforms all methods in average accuracy with both settings. As illustrated in Fig.~\ref{subfig:comparison-10steps(a)} and Fig.~\ref{subfig:comparison-50steps(a)}, the accuracies of the proposed method are 1.73\% and 3.71\% higher than those of the runner-up after learning all tasks in the settings of $T$=10 and $T$=50, respectively, demonstrating its stronger resistance to catastrophic forgetting.

\begin{figure*}[t]
\centering
\subfloat[\scriptsize{Different learning orders} \label{subfig:addition_exp(a)}]{\includegraphics[width=1.7in]{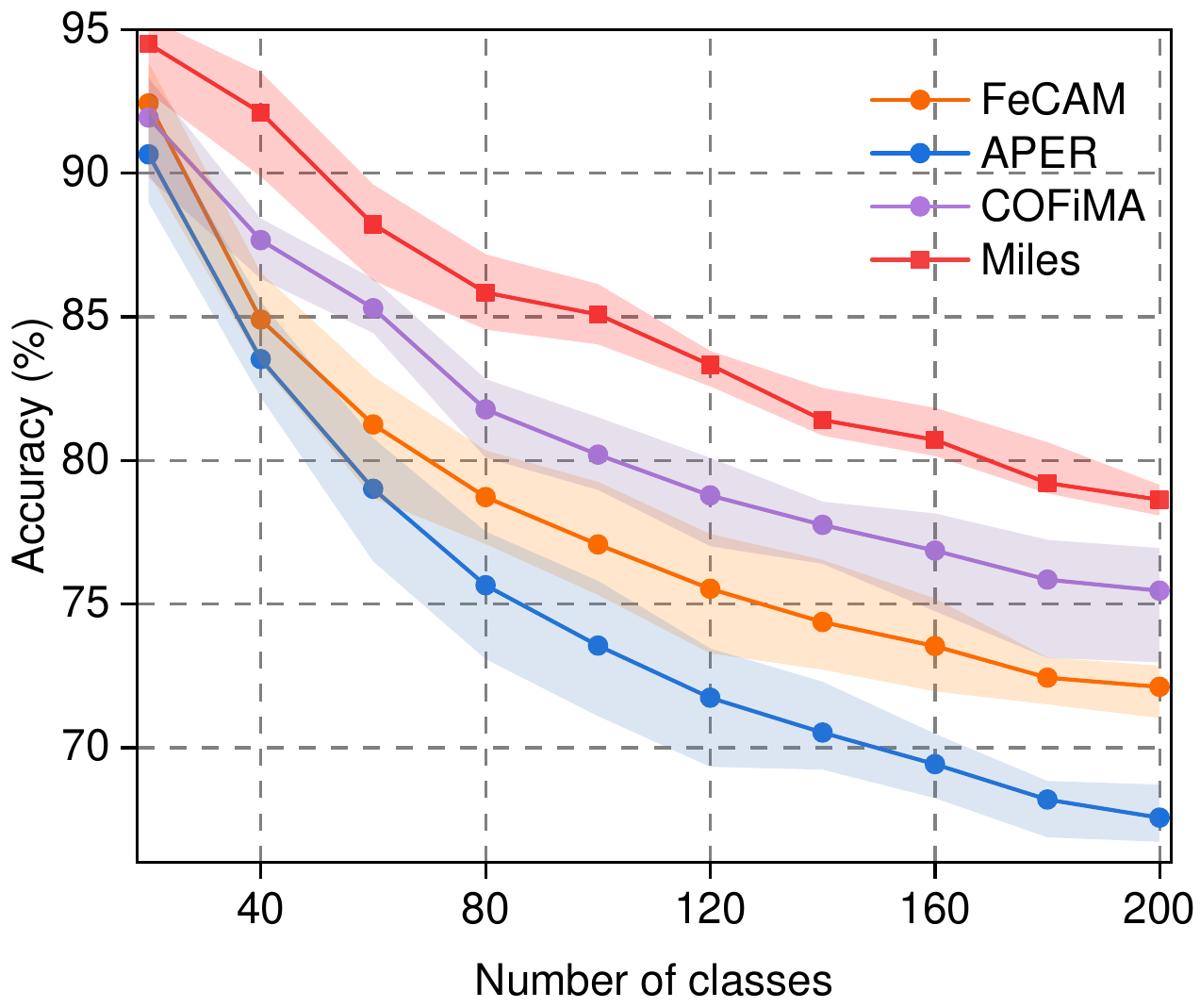}}
\hfill
\subfloat[\scriptsize{Sensitivity experiments} \label{subfig:addition_exp(b)}]{\includegraphics[width=1.85in]{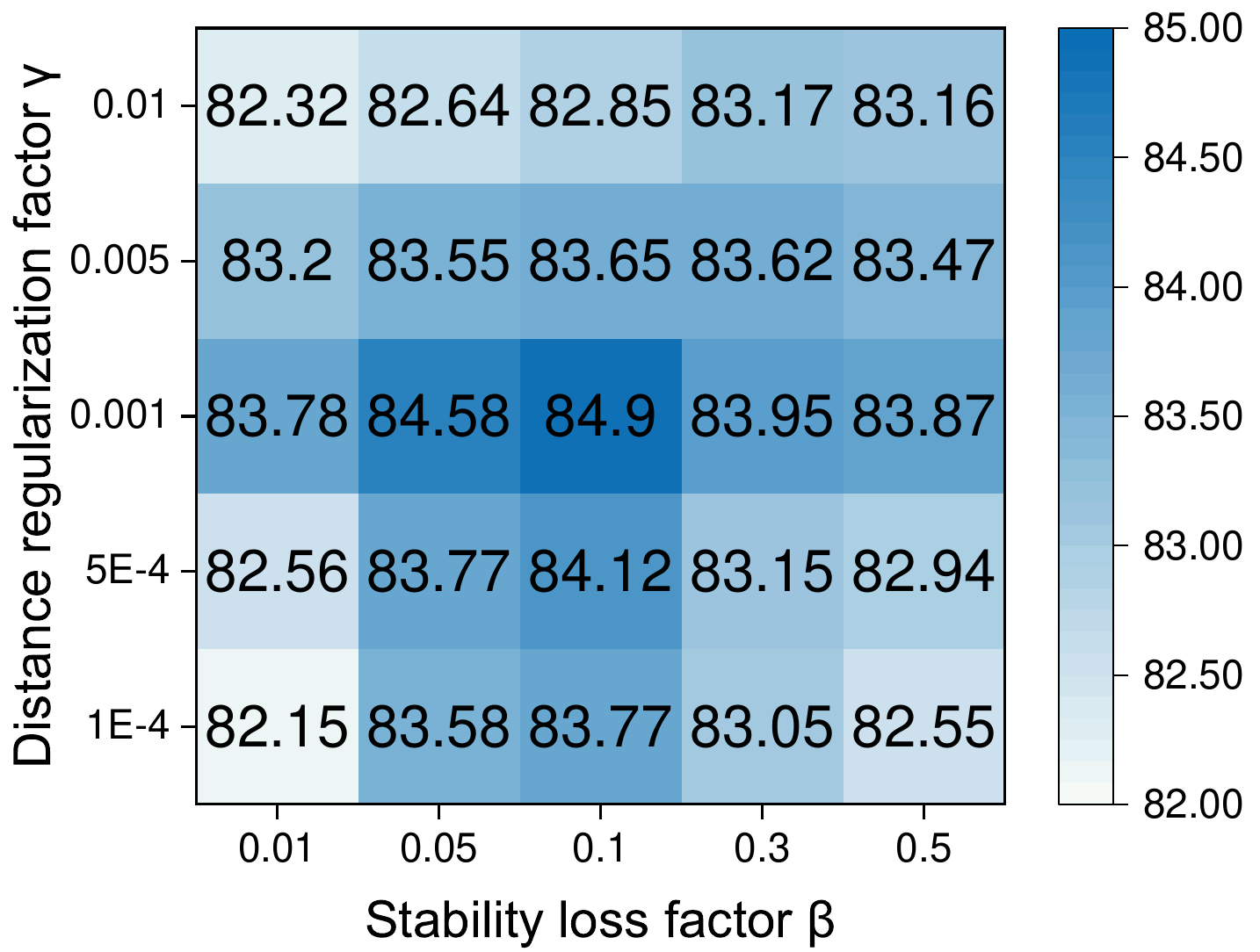}}
\hfill
\subfloat[\scriptsize{Model parameters} \label{subfig:addition_exp(c)}]{\includegraphics[width=1.7in]{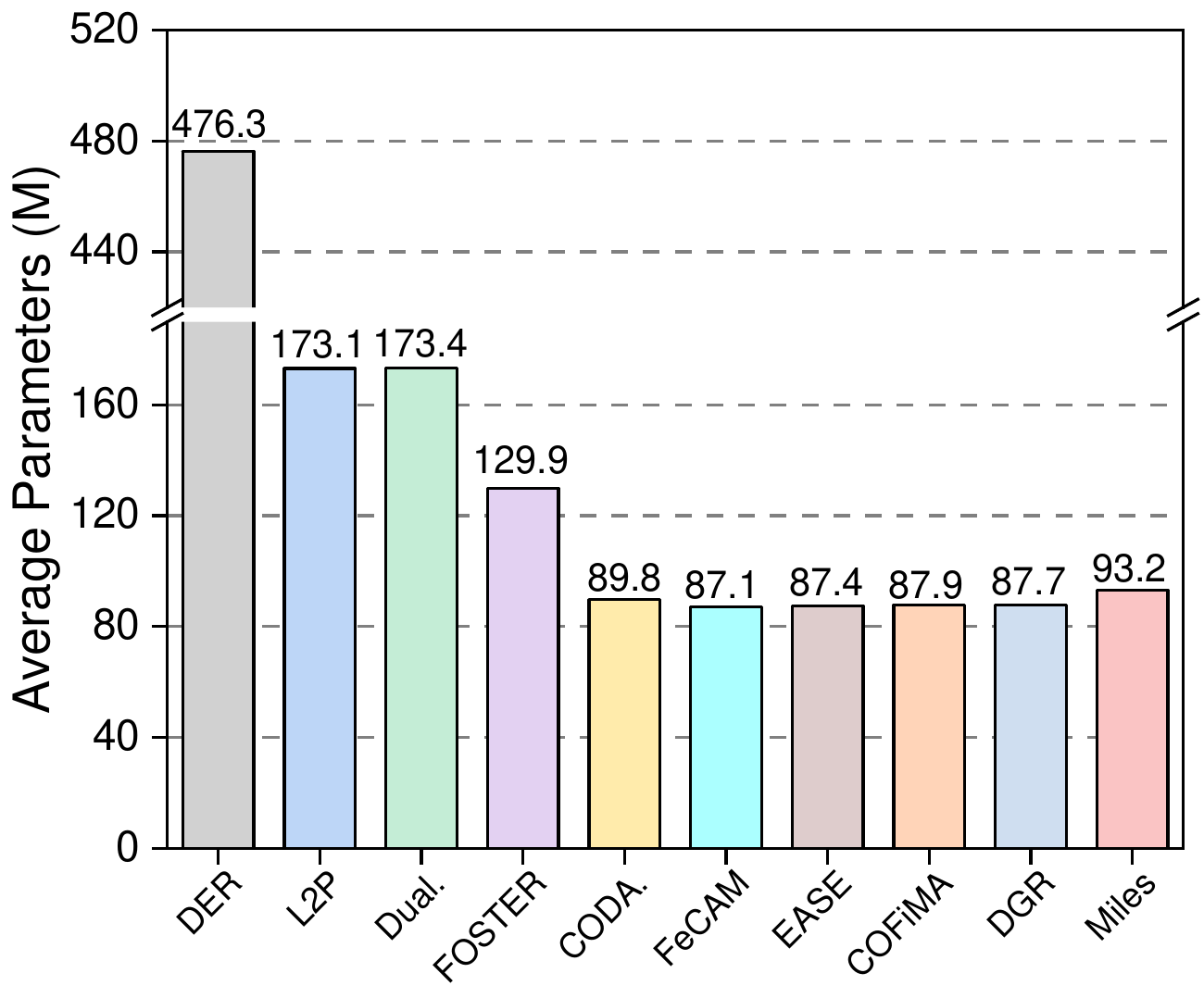}}
\hfill
\subfloat[\scriptsize{Computation complexity} \label{subfig:addition_exp(d)}]{\includegraphics[width=1.7in]{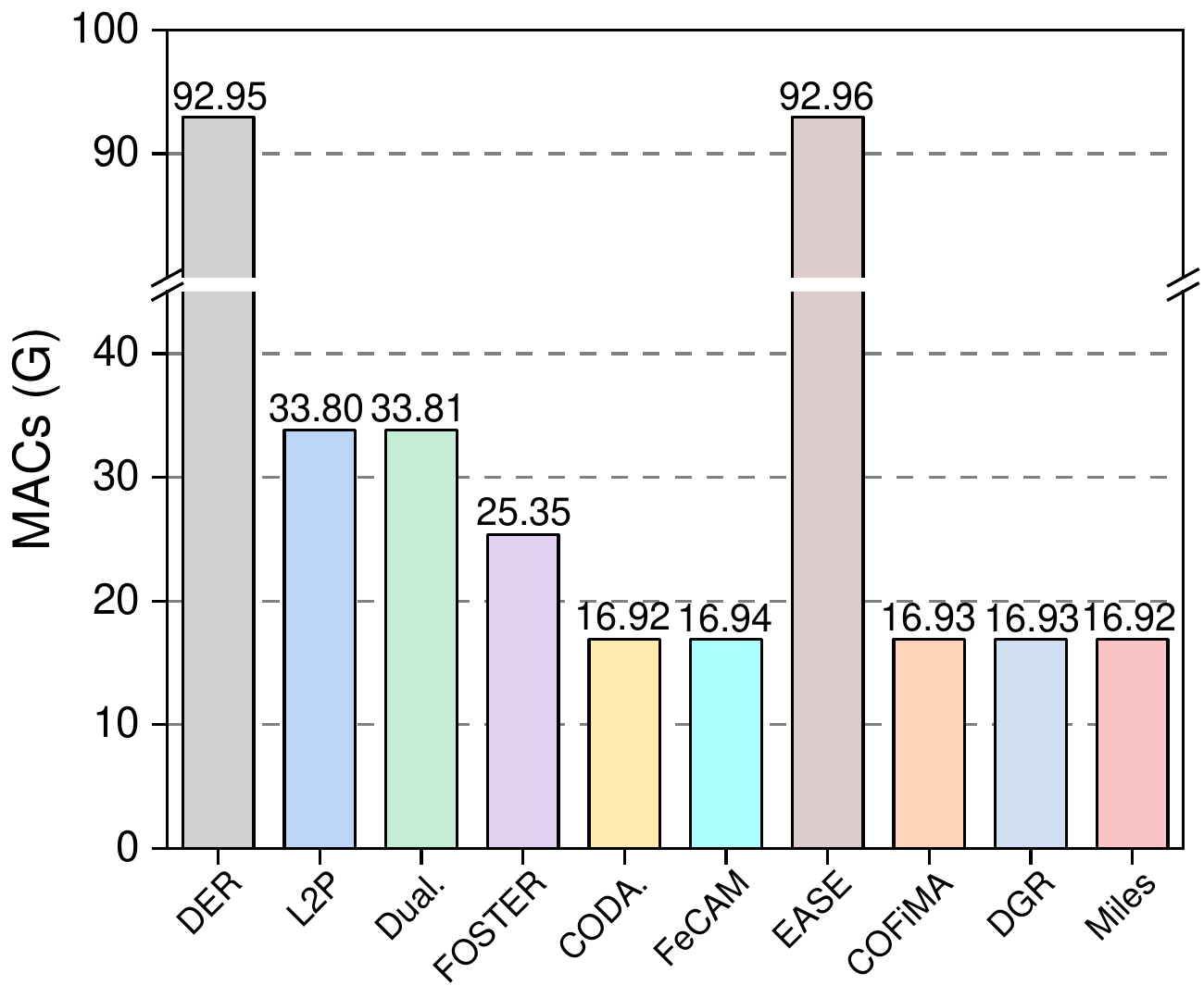}}
\caption{Additional experiments. (a) \textbf{\textsc{Miles} with different learning order}: Different learning orders are applied for \textsc{Miles} in the ImageNet-R dataset. The learning orders are determined by random seeds of $\left\{ {1994,1995,1996,1997,1998} \right\}$. (b) \textbf{Hyperparameter sensitivity experiment}: We validate \textsc{Miles} with different combinations of hyper-parameters in the ImageNet-R datasets and report the average accuracy. (c) \textbf{Analysis of model parameters in various CIL methods}: The average number of model parameters of each task is reported for different CIL methods in the setting of 10 steps. (d) \textbf{Computation complexity}: Average GMACs of each task is report for each CIL method in the setting of 10 steps.} \label{fig:addition_exp}
\end{figure*}

\paragraph{CUB} As illustrated in Tab.~\ref{tab:cmp1}, the average accuracies of \textsc{Miles} in the settings of $T$=10 and 50 are 93.02\% and 92.86\% for the CUB dataset, achieving the best performance in all CIL methods. Moreover, as illustrated in Fig.~\ref{subfig:comparison-10steps(b)} and Fig.~\ref{subfig:comparison-50steps(b)}, the accuracies of the proposed method are 3.04\% and 2.11\% higher than those of the runner-ups after learning all tasks in the settings of $T$=10 and 50, respectively, demonstrating that the proposed method outperforms other CIL methods by a significant margin.

\paragraph{Omnibenchmark} As illustrated in Tab.~\ref{tab:cmp1}, the average accuracies of \textsc{Miles} in the settings of $T$=10 and 50 are 85.02\% and 84.84\% for the Omnibenchmark dataset, which outperform the runner-up by 1.88\% and 2.14\%, respectively. As illustrated in Fig.~\ref{subfig:comparison-10steps(c)} and Fig.~\ref{subfig:comparison-50steps(c)}, the accuracies of the proposed method are 2.42\% and 2.85\% higher than those of the runner-up after learning all tasks in the settings of $T$=10 and 50, demonstrating the superior performance of the proposed method.

\paragraph{ImageNet-A} As illustrated in Tab.~\ref{tab:cmp2}, the average accuracies of \textsc{Miles} in the settings of $T$=10 and 50 are 62.33\% and 62.67\% for the ImageNet-A dataset. As illustrated in Fig.~\ref{subfig:comparison-10steps(d)} and Fig.~\ref{subfig:comparison-50steps(d)}, the accuracies of the proposed method are 2.83\% and 4.28\% higher than those of the runner-ups after learning all tasks in the settings of $T$=10 and 50, respectively, demonstrating that the proposed method outperforms other CIL methods in this dataset.

\paragraph{ImageNet-R} As illustrated in Tab.~\ref{tab:cmp2}, the average accuracies of \textsc{Miles} in the settings of $T$=10 and 50 are 84.90\% and 81.90\% for the ImageNet-R dataset. As illustrated in Fig.~\ref{subfig:comparison-10steps(e)} and Fig.~\ref{subfig:comparison-50steps(e)}, the accuracies of the proposed method are 2.20\% and 3.54\% higher than those of the runner-up after learning all tasks in the settings of $T$=10 and 50, also indicating the superiority of the proposed method.

\paragraph{FOOD} As illustrated in Tab.~\ref{tab:cmp2}, the average accuracies of \textsc{Miles} in the settings of $T$=10 and 50 are 93.23\% and 90.75\% in the FOOD-101 dataset, which outperform the runner-up by 2.08\% and 1.54\%, respectively. As illustrated in Fig.~\ref{subfig:comparison-10steps(f)} and Fig.~\ref{subfig:comparison-50steps(f)}, the accuracies of the proposed method are 1.85\% and 2.80\% higher than those of the runner-up after learning all tasks in the settings of $T$=10 and 50, validating the superior performance of the proposed method.

\begin{table*}[h]
\centering
\caption{Results of ablation study.}
\resizebox{1.0\linewidth}{!}{
\small
\begin{tabular}{lcc|cc|cc|cc|cc|cc}
    \toprule
    \multirow{3}{*}{\large{\textbf{Strategy}}} & \multicolumn{2}{c|}{\small{CIFAR}} & \multicolumn{2}{c|}{\small{CUB}} & \multicolumn{2}{c|}{\small{Omnibenchmark}} & \multicolumn{2}{c|}{\small{ImageNet-A}} & \multicolumn{2}{c|}{\small{ImageNet-R}} & \multicolumn{2}{c}{\small{FOOD}}\\
    \cmidrule(lr){2-13}
     & $\overline{A}\uparrow$ & $F\downarrow$ & $\overline{A}\uparrow$ & $F\downarrow$ & $\overline{A}\uparrow$ & $F\downarrow$ & $\overline{A}\uparrow$ & $F\downarrow$ & $\overline{A}\uparrow$ & $F\downarrow$ & $\overline{A}\uparrow$ & $F\downarrow$\\
    \midrule
Baseline & $87.92$ & $10.46$ & $87.62$ & $7.14$ & $79.85$ & $9.67$ & $58.98$ & $4.56$ & $71.80$ & $21.22$ & $86.92$ & $9.21$\\
	\midrule[0.5pt]
w/ SPP-MLP & $89.34 \scriptstyle \pm 0.05$ & $7.95 \scriptstyle \pm 0.06$ & $88.15 \scriptstyle \pm 0.10$ & $7.00 \scriptstyle \pm 0.09$ & $81.06 \scriptstyle \pm 0.14$ & $7.65 \scriptstyle \pm 0.12$ & $59.42 \scriptstyle \pm 0.28$ & $4.28 \scriptstyle \pm 0.25$ & $77.93 \scriptstyle \pm 0.10$ & $14.55 \scriptstyle \pm 0.08$ & $88.47 \scriptstyle \pm 0.05$ & $7.63 \scriptstyle \pm 0.07$\\
w/ SPP-Bottleneck & $90.06 \scriptstyle \pm 0.08$ & $7.18 \scriptstyle \pm 0.08$ & $89.35 \scriptstyle \pm 0.14$ & $6.55 \scriptstyle \pm 0.12$ & $81.93 \scriptstyle \pm 0.08$ & $6.88 \scriptstyle \pm 0.10$ & $60.15 \scriptstyle \pm 0.20$ & $3.79 \scriptstyle \pm 0.22$ & $80.96 \scriptstyle \pm 0.15$ & $10.69 \scriptstyle \pm 0.13$ & $89.22 \scriptstyle \pm 0.06$ & $6.44 \scriptstyle \pm 0.08$\\
w/ SPP-Full & $91.68 \scriptstyle \pm 0.07$ & $6.24 \scriptstyle \pm 0.10$ & $90.21 \scriptstyle \pm 0.06$ & $6.43 \scriptstyle \pm 0.08$ & $82.47 \scriptstyle \pm 0.12$ & $6.25 \scriptstyle \pm 0.14$ & $61.06 \scriptstyle \pm 0.24$ & $3.52 \scriptstyle \pm 0.19$ & $82.63 \scriptstyle \pm 0.15$ & $9.67 \scriptstyle \pm 0.10$ & $90.58 \scriptstyle \pm 0.07$ & $4.95 \scriptstyle \pm 0.08$\\
	\midrule[0.5pt]
w/ SPP + SRE & $92.88 \scriptstyle \pm 0.12$ & $3.85 \scriptstyle \pm 0.08$ & $91.62 \scriptstyle \pm 0.11$ & $5.48 \scriptstyle \pm 0.15$ & $83.92 \scriptstyle \pm 0.15$ & $4.86 \scriptstyle \pm 0.10$ & $62.00 \scriptstyle \pm 0.22$ & $3.05 \scriptstyle \pm 0.16$ & $83.70 \scriptstyle \pm 0.17$ & $6.88 \scriptstyle \pm 0.12$ & $92.06 \scriptstyle \pm 0.10$ & $4.49 \scriptstyle \pm 0.04$\\
w/ SPP + SRE + DR & $93.89 \scriptstyle \pm 0.08$ & $3.20 \scriptstyle \pm 0.10$ & $93.13 \scriptstyle \pm 0.08$ & $4.55 \scriptstyle \pm 0.06$ & $84.96 \scriptstyle \pm 0.08$ & $4.12 \scriptstyle \pm 0.08$ & $62.41 \scriptstyle \pm 0.28$ & $2.87 \scriptstyle \pm 0.19$ & $84.79 \scriptstyle \pm 0.23$ & $5.76 \scriptstyle \pm 0.21$ & $93.28 \scriptstyle \pm 0.08$ & $3.92 \scriptstyle \pm 0.10$\\
    \bottomrule
\end{tabular}
}
\label{tab:ablation}
\end{table*}

\begin{table}[t]
\caption{Sensitivity experiments on hyperparameters.}\label{tab:sensitive_rd}
\centering‌
\begin{tabular}{c|cccc}
\toprule
ImageNet-R $T=10$ & $d_3=4$ & $d_3=16$ & $d_3=48$ & $d_3=96$\\
\midrule[0.5pt]
$r=2$ & 84.52 & 84.64 & 84.89 & 84.90\\
$r=4$ & 83.12 & 84.75 & 84.92 & 85.33\\
$r=8$ & 83.47 & 83.77 & 83.98 & 84.25\\
\bottomrule
\end{tabular}
\end{table}

\subsection{Ablation study} \label{subsec:ablation}
To verify the effectiveness of each component in \textsc{Miles}, we conduct ablation studies using the setting of $T$=10 on all six datasets. To eliminate randomness, each experiment was repeated three times, and the \textbf{mean} and \textbf{standard deviation} are reported. As shown in Tab.~\ref{tab:ablation}, Row 3 shows the results of the baseline method, which merely adopts the prototype for classification without training. Row 4 shows the results of applying the SPP strategy with only a MLP layer to construct a subspace for learning a new task. Row 5 shows the results of applying the SPP strategy with bottleneck structure to construct a subspace for learning a new task. Row 6 shows the results of applying the complete SPP strategy to construct a subspace for learning a new task. \textbf{Comparing the lines 3 to 6, the progressive structure built by the SPP module performs better than the basic MLP layers and the general bottleneck structure}. Row 7 shows the results of adding the SPP and SRE modules to strengthen the representative capacity of the backbone network. Row 8 shows the results of adding the DR strategy to balance the distance between features and prototypes in the feature subspace. \textbf{The comparison between lines 7 and 8 shows that the DR strategy is effective on most benchmarks. Moreover, based on the standard deviation, it can be determined that its improvement to the model is not caused by randomness. After incrementally adding three strategies to the baseline method, the model performance is steadily improved, demonstrating the effectiveness of each component.}

\subsection{Additional experiments and visualizations} \label{subsec:vis}
\paragraph{Different learning orders} As illustrated in Fig.~\ref{subfig:addition_exp(a)}, we validate the \textsc{Miles} with 5 learning orders, which are determined by a random seed selected from $\left\{ {1994,1995,1996,1997,1998} \right\}$. The results show that \textsc{Miles} exhibits only a small fluctuation across various learning orders, demonstrating its robustness to order alterations.
\paragraph{Parameter robustness} There are two hyper-parameters in \textsc{Miles}, i.e., the stability loss factor $\beta$ and the distance regularization factor $\gamma$ defined in Eq.~\eqref{eq:26}. We conduct hyper-parameter sensitivity experiment to investigate the parameter robustness by selecting different combinations of the two hyper-parameters on ImageNet-R under the $T$=10 setting. Specifically, we choose $\beta$ from $\left\{ {0.01,0.05,0.1,0.3,0.5} \right\}$ and $\gamma$ from $\left\{ {1 \times {{10}^{ - 4}},5 \times {{10}^{ - 4}},0.001,0.005,0.01} \right\}$, and report the average performance in Fig.~\ref{subfig:addition_exp(b)}. It is observed that the performance remains relatively stable, and we suggest $\beta = 0.1$, $\gamma = 0.001$ as default for other datasets. Additionally, we provide further sensitivity experiments for the hyperparameters $r$ and $d_3$ in Tab.~\ref{tab:sensitive_rd} with the setting of $T$=10 in ImageNet-R. Specifically, we select $r$ from $\left\{ {2,4,8} \right\}$ and $d_3$ from $\left\{ {4,16,48,96} \right\}$, and report the average accuracy. It is observed that the performance remains stable, and we suggest $r = 4$ and ${d_3} = 16$ to achieve the tradeoff between efficiency and performance.
\paragraph{Computational overhead} We compare the number of model parameters and computational complexity for all CIL approaches. As shown in Fig.~\ref{subfig:addition_exp(c)}, we report the average number of model parameters in the setting of $T$=10. It demonstrates that \textsc{Miles} maintains a relatively small number of model parameters. Furthermore, assuming FP32 storage per parameter, then the proposed method has a memory consumption of approximately 373 MB, comparable to that of the smallest model, FeCAM~(348 MB). In particular, the memory consumption for the exemplar set is not considered for the memory budget. Therefore, exemplar-based methods such as FOSTER and DGR actually require more memory consumption. Furthermore, we report the \textit{Multiply-Accumulate Operations}~(MACs) for reasoning a 224$\times$224 colored image. As shown in Fig.~\ref{subfig:addition_exp(d)}, the proposed method maintains a minimal computational cost among the comparative CIL methods. In particular, although EASE has only 87.4M parameters, each task branch requires a separate forward pass through the backbone network during inference. Consequently, its computational cost becomes comparable to that of the network expansion approach DER.

\begin{table*}[t]
\caption{Intra-class/Inter-class Distance Analysis under Random/Pre-trained Weights of Multiple Models}\label{tab:dis_analysis}
\centering
‌\small
\begin{tabular*}{\textwidth}{@{\extracolsep{\fill}}lcccccccc}
\toprule
\multirow{4}{*}{{\bf Backbone}} & \multicolumn{4}{c}{CIFAR} & \multicolumn{4}{c}{CUB}\\
  \cmidrule(lr){2-9}
  & \multicolumn{2}{c}{Random} & \multicolumn{2}{c}{Pretrain} & \multicolumn{2}{c}{Random} & \multicolumn{2}{c}{Pretrain}\\
  \cmidrule(lr){2-3} \cmidrule(lr){4-5} \cmidrule(lr){6-7} \cmidrule(lr){8-9}
 & Intra & Inter & Intra & Inter & Intra & Inter & Intra & Inter\\
 \midrule
ResNet50 & $4.370 \scriptstyle \pm 0.945$ & $4.976 \scriptstyle \pm 1.327$ & $0.092 \scriptstyle \pm 0.012$ & $0.117 \scriptstyle \pm 0.010$ & $1.916 \scriptstyle \pm 0.309$ & $2.059 \scriptstyle \pm 0.387$ & $0.044 \scriptstyle \pm 0.009$ & $0.062 \scriptstyle \pm 0.009$\\
ViT-B/16 & $0.552 \scriptstyle \pm 0.064$ & $0.658 \scriptstyle \pm 0.091$ & $0.113 \scriptstyle \pm 0.013$ & $0.156 \scriptstyle \pm 0.009$ & $0.569 \scriptstyle \pm 0.067$ & $0.639 \scriptstyle \pm 0.091$ & $0.068 \scriptstyle \pm 0.016$ & $0.153 \scriptstyle \pm 0.014$\\
Swin-B   & $0.628 \scriptstyle \pm 0.077$ & $0.720 \scriptstyle \pm 0.071$ & $0.182 \scriptstyle \pm 0.037$ & $0.215 \scriptstyle \pm 0.032$ & $0.666 \scriptstyle \pm 0.059$ & $0.734 \scriptstyle \pm 0.067$ & $0.071 \scriptstyle \pm 0.018$ & $0.114 \scriptstyle \pm 0.023$\\
ConvNexT & $0.697 \scriptstyle \pm 0.096$ & $0.841 \scriptstyle \pm 0.096$ & $0.128 \scriptstyle \pm 0.018$ & $0.175 \scriptstyle \pm 0.015$ & $0.718 \scriptstyle \pm 0.073$ & $0.821 \scriptstyle \pm 0.095$ & $0.077 \scriptstyle \pm 0.020$ & $0.131 \scriptstyle \pm 0.022$\\
\midrule[0.5pt]
{\bf Backbone} & \multicolumn{4}{c}{Omnibenchmark} & \multicolumn{4}{c}{ImageNet-A}\\
\midrule[0.5pt]
ResNet50 & $2.478 \scriptstyle \pm 0.580$ & $2.980 \scriptstyle \pm 0.885$ & $0.061 \scriptstyle \pm 0.024$ & $0.090 \scriptstyle \pm 0.027$ & $2.067 \scriptstyle \pm 0.728$ & $2.261 \scriptstyle \pm 0.816$ & $0.129 \scriptstyle \pm 0.019$ & $0.147 \scriptstyle \pm 0.016$\\
ViT-B/16 & $0.537 \scriptstyle \pm 0.079$ & $0.645 \scriptstyle \pm 0.094$ & $0.090 \scriptstyle \pm 0.016$ & $0.150 \scriptstyle \pm 0.013$ & $0.571 \scriptstyle \pm 0.097$ & $0.604 \scriptstyle \pm 0.091$ & $0.142 \scriptstyle \pm 0.011$ & $0.154 \scriptstyle \pm 0.009$\\
Swin-B   & $0.621 \scriptstyle \pm 0.106$ & $0.724 \scriptstyle \pm 0.106$ & $0.096 \scriptstyle \pm 0.040$ & $0.151 \scriptstyle \pm 0.037$ & $0.699 \scriptstyle \pm 0.080$ & $0.728 \scriptstyle \pm 0.069$ & $0.200 \scriptstyle \pm 0.077$ & $0.220 \scriptstyle \pm 0.076$\\
ConvNexT & $0.668 \scriptstyle \pm 0.121$ & $0.820 \scriptstyle \pm 0.134$ & $0.085 \scriptstyle \pm 0.036$ & $0.153 \scriptstyle \pm 0.028$ & $0.770 \scriptstyle \pm 0.117$ & $0.822 \scriptstyle \pm 0.096$ & $0.166 \scriptstyle \pm 0.023$ & $0.184 \scriptstyle \pm 0.019$\\
\midrule[0.5pt]
{\bf Backbone} & \multicolumn{4}{c}{ImageNet-R} & \multicolumn{4}{c}{FOOD}\\
\midrule[0.5pt]
ResNet50 & $3.344 \scriptstyle \pm 0.637$ & $3.567 \scriptstyle \pm 0.704$ & $0.110 \scriptstyle \pm 0.011$ & $0.125 \scriptstyle \pm 0.008$ & $2.103 \scriptstyle \pm 0.226$ & $2.380 \scriptstyle \pm 0.475$ & $0.064 \scriptstyle \pm 0.012$ & $0.075 \scriptstyle \pm 0.011$\\
ViT-B/16 & $0.610 \scriptstyle \pm 0.052$ & $0.639 \scriptstyle \pm 0.049$ & $0.132 \scriptstyle \pm 0.010$ & $0.150 \scriptstyle \pm 0.007$ & $0.424 \scriptstyle \pm 0.041$ & $0.452 \scriptstyle \pm 0.040$ & $0.098 \scriptstyle \pm 0.016$ & $0.150 \scriptstyle \pm 0.011$\\
Swin-B   & $0.603 \scriptstyle \pm 0.073$ & $0.621 \scriptstyle \pm 0.063$ & $0.151 \scriptstyle \pm 0.031$ & $0.176 \scriptstyle \pm 0.030$ & $0.670 \scriptstyle \pm 0.016$ & $0.685 \scriptstyle \pm 0.016$ & $0.144 \scriptstyle \pm 0.020$ & $0.172 \scriptstyle \pm 0.019$\\
ConvNexT & $0.667 \scriptstyle \pm 0.101$ & $0.699 \scriptstyle \pm 0.092$ & $0.144 \scriptstyle \pm 0.020$ & $0.172 \scriptstyle \pm 0.012$ & $0.641 \scriptstyle \pm 0.070$ & $0.688 \scriptstyle \pm 0.065$ & $0.106 \scriptstyle \pm 0.018$ & $0.149 \scriptstyle \pm 0.017$\\
\bottomrule
\end{tabular*}
\end{table*}

\begin{figure*}[t]
\centering
\subfloat[\scriptsize{Original feature space}\label{subfig:vis1(a)}]{\includegraphics[width=1.7in]{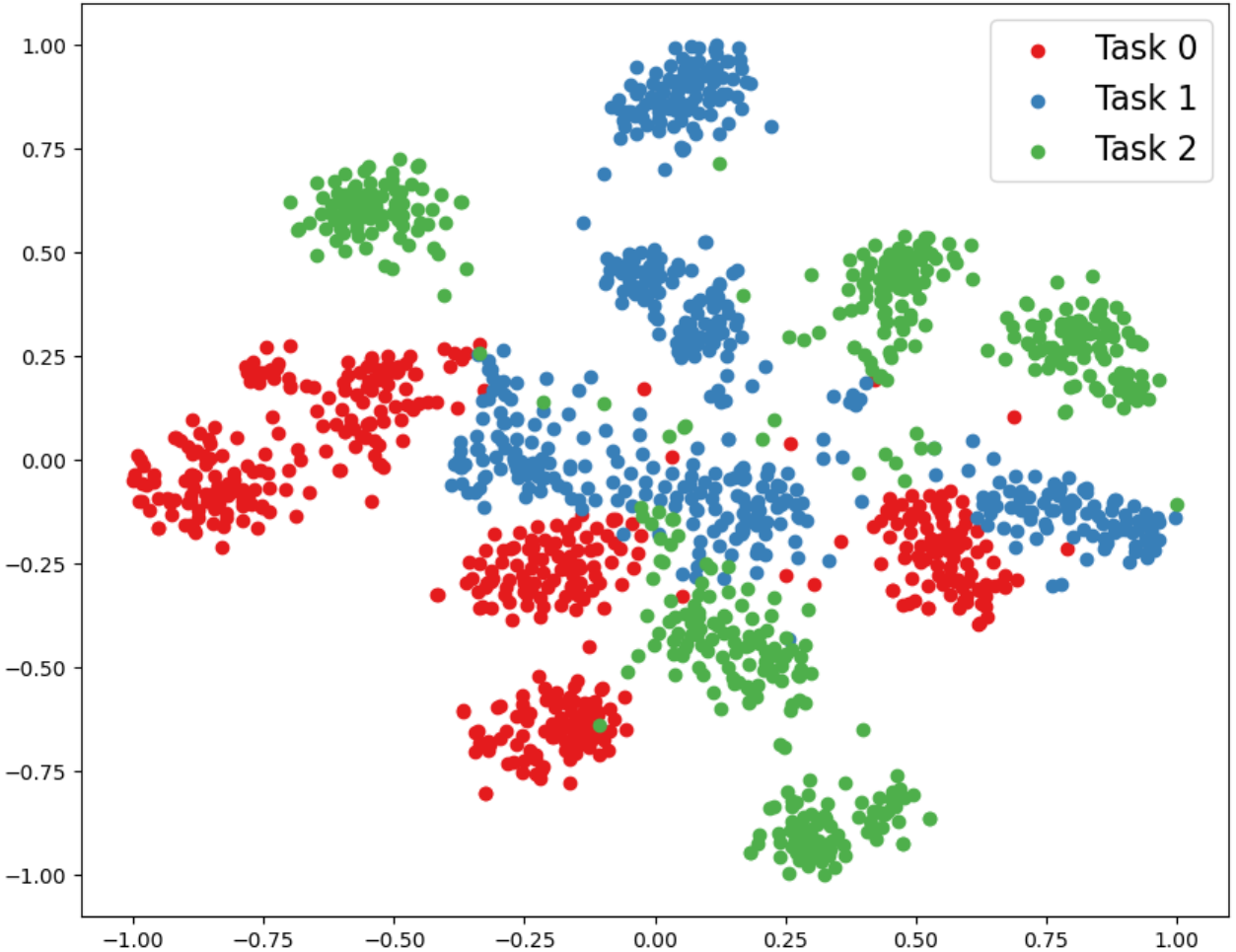}}
\hfill
\subfloat[\scriptsize{Feature subspace of task 0}\label{subfig:vis1(b)}]{\includegraphics[width=1.7in]{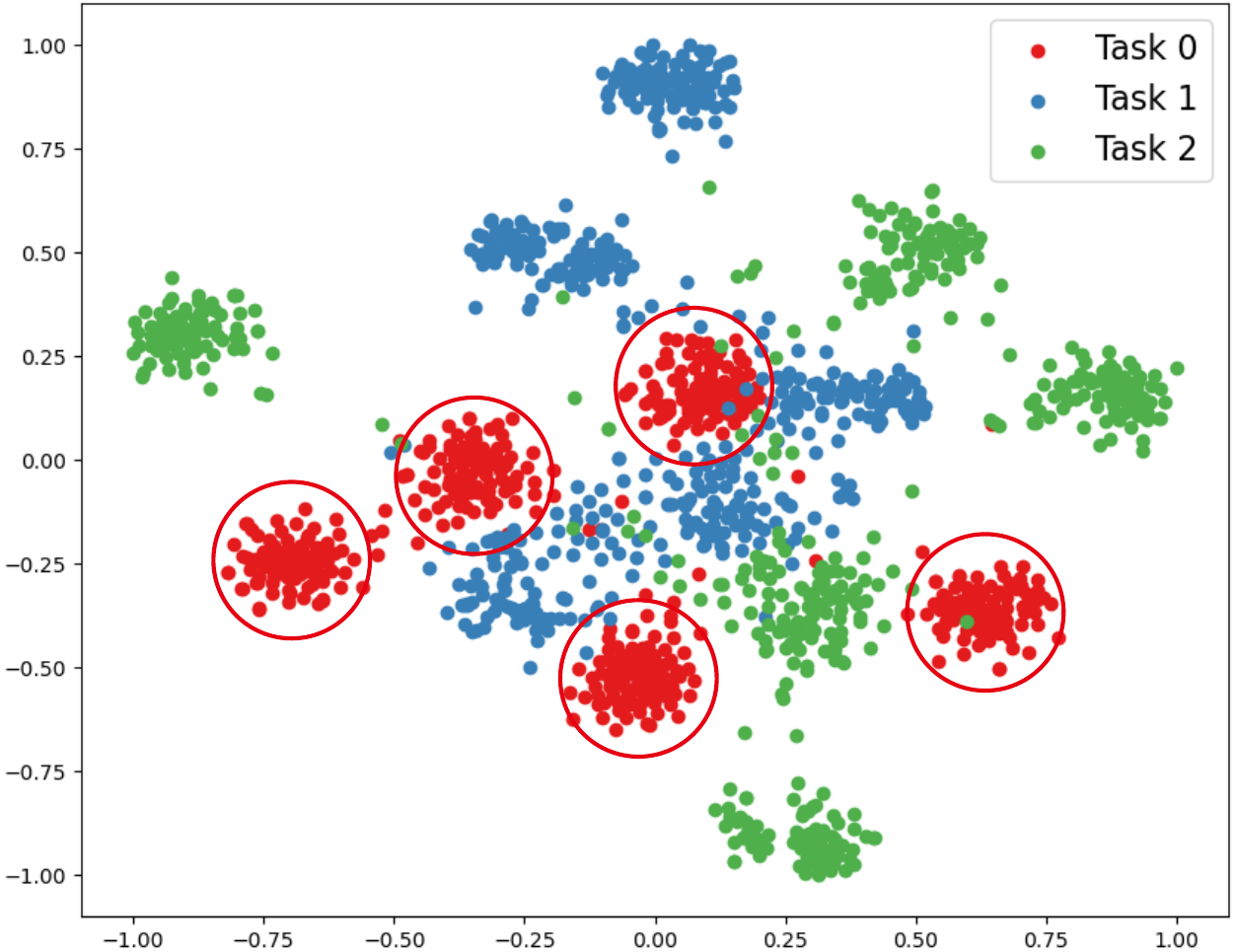}}
\hfill
\subfloat[\scriptsize{Feature subspace of task 1}\label{subfig:vis1(c)}]{\includegraphics[width=1.7in]{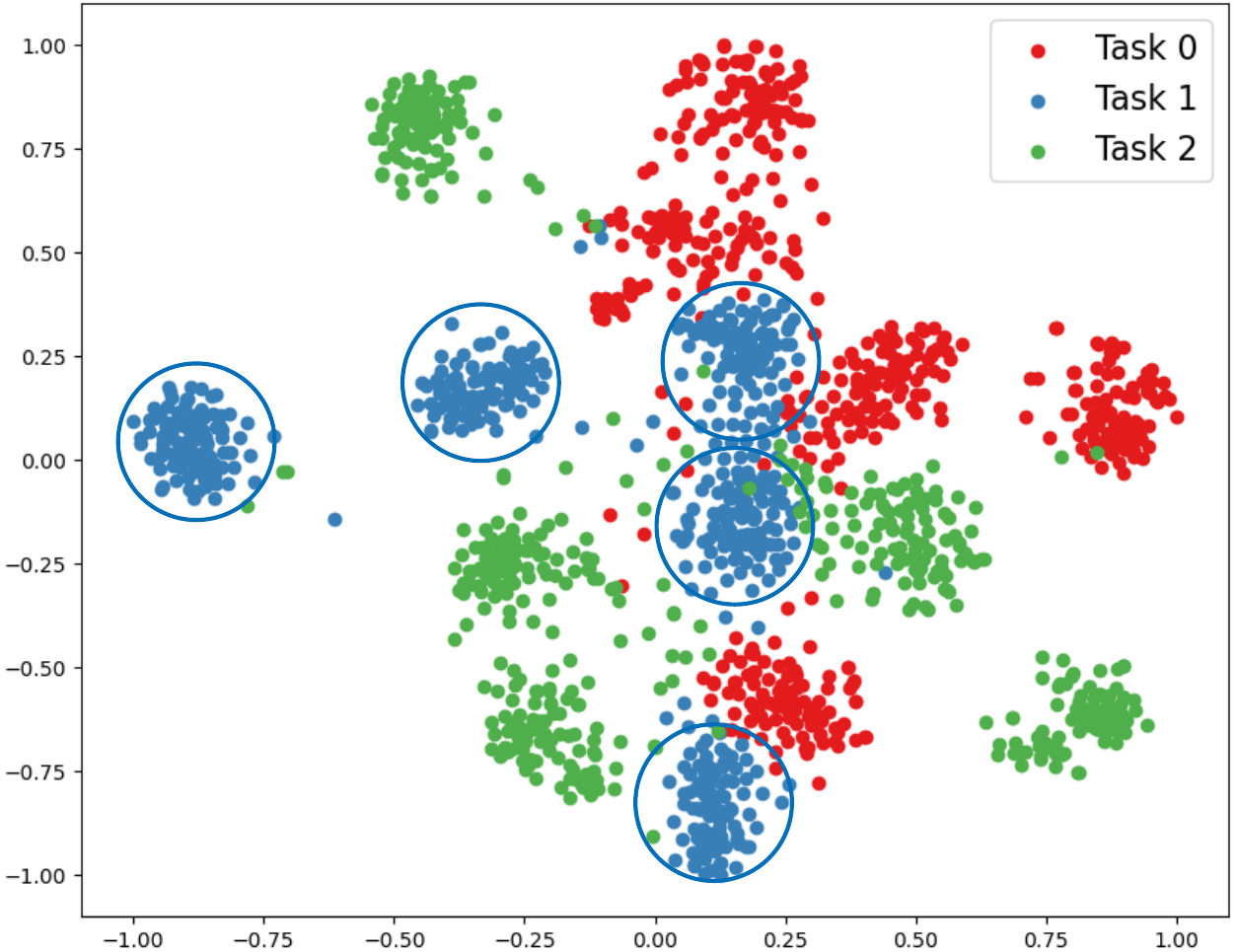}}
\hfill
\subfloat[\scriptsize{Feature subspace of task 2}\label{subfig:vis1(d)}]{\includegraphics[width=1.7in]{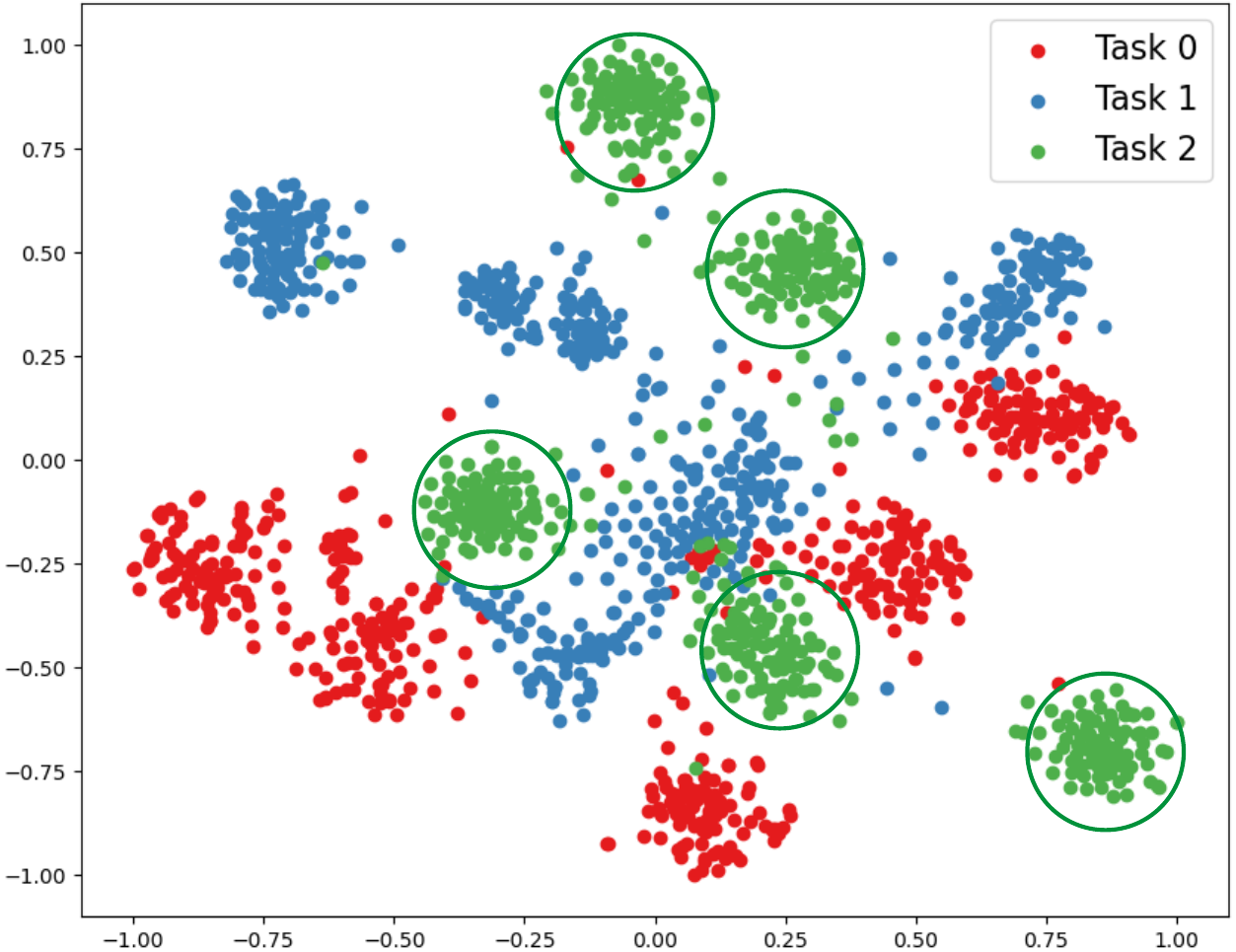}}
\caption{T-SNE visualization in original feature space and feature subspaces. The model is incrementally trained on three tasks within the CIFAR-100 dataset, each comprising five categories. Subsequently, T-SNE visualization is applied to the original features and all subspace projection features.} \label{fig:vis1}
\end{figure*}

\begin{figure*}[t]
\centering
\subfloat[\scriptsize{Learning 10 categories}\label{subfig:vis2(a)}]{\includegraphics[width=3.4in]{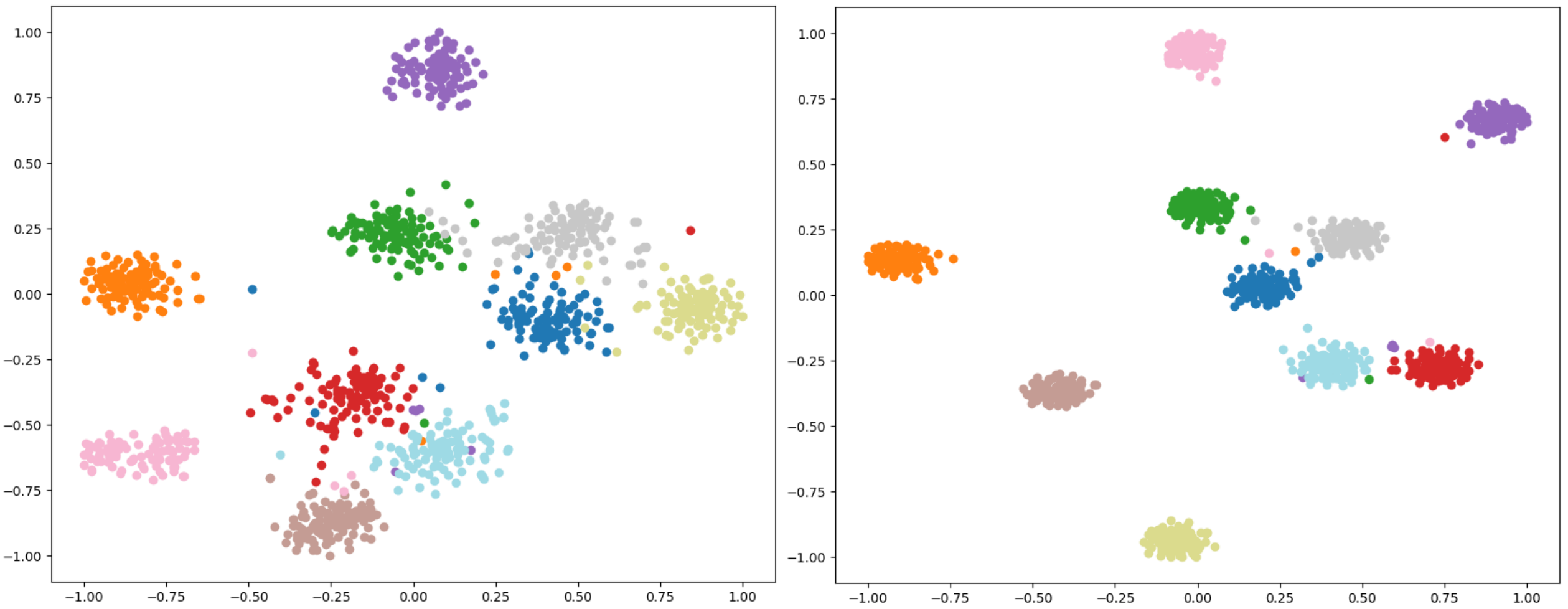}}
\hfill
\subfloat[\scriptsize{Learning 20 categories}\label{subfig:vis2(b)}]{\includegraphics[width=3.4in]{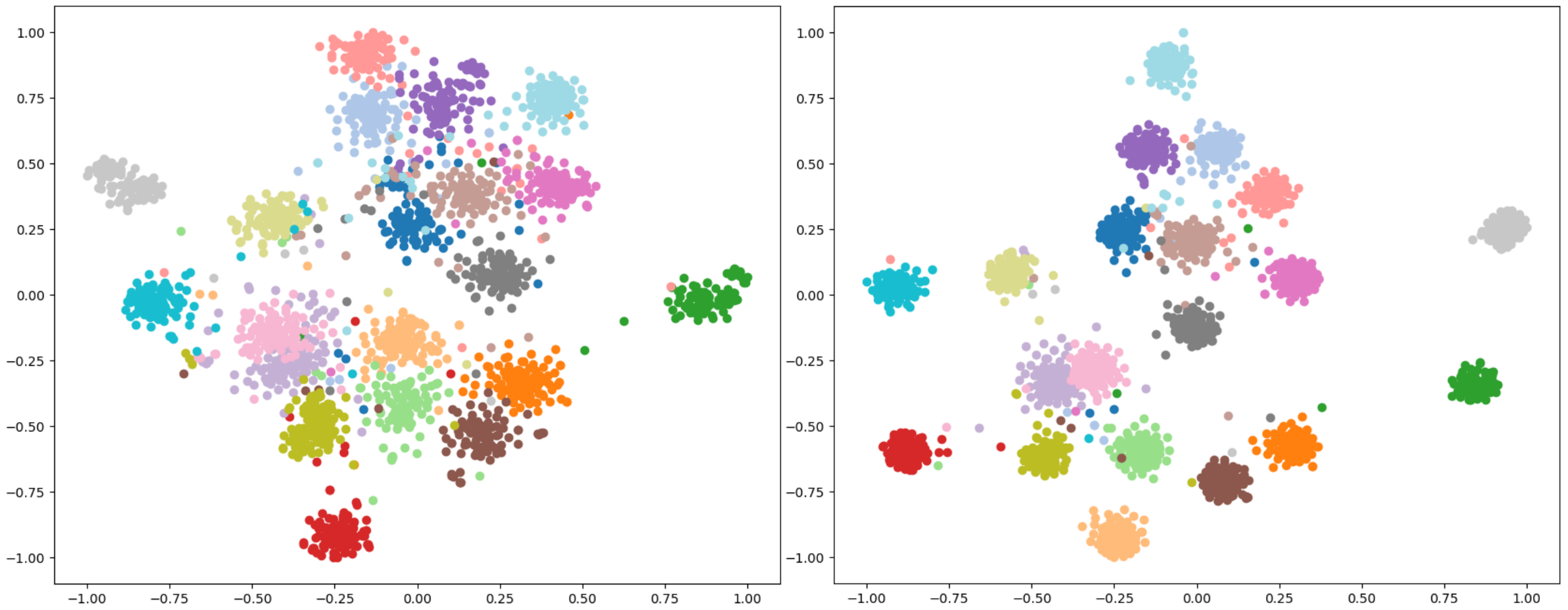}}
\hfill
\\
\subfloat[\scriptsize{Learning 50 categories}\label{subfig:vis2(c)}]{\includegraphics[width=3.4in]{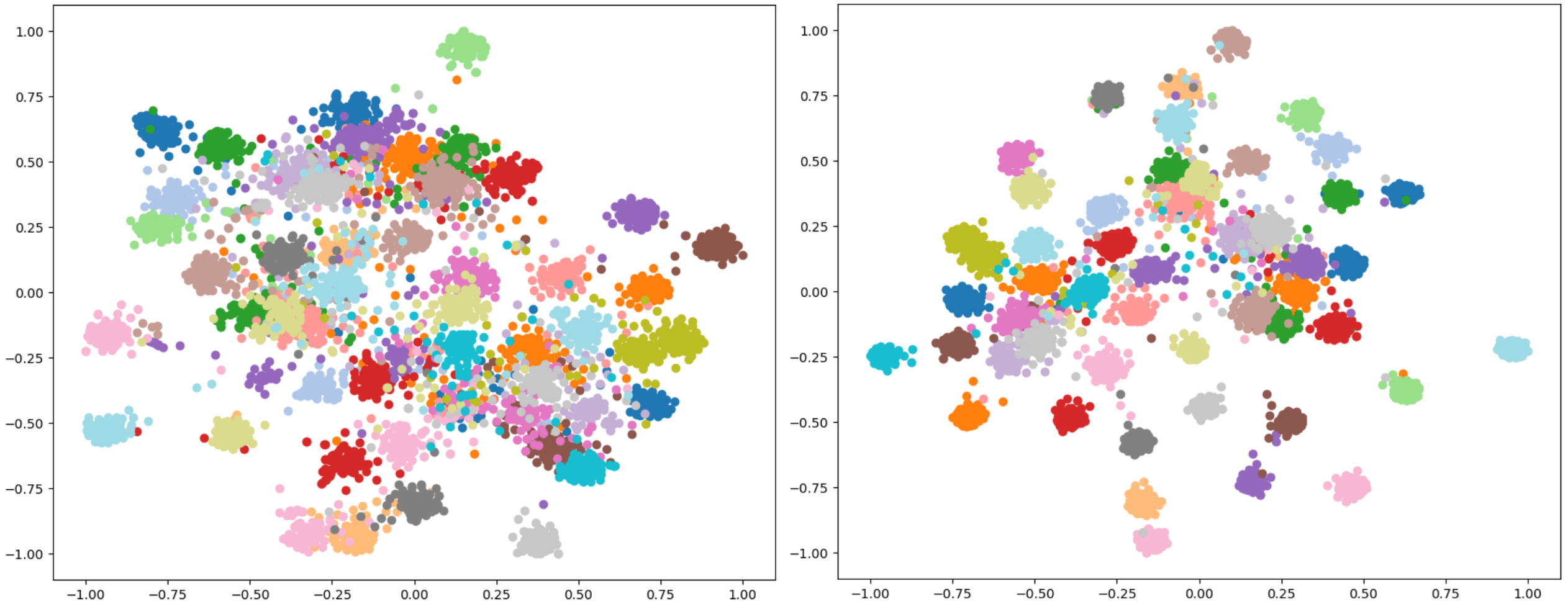}}
\hfill
\subfloat[\scriptsize{Learning 100 categories}\label{subfig:vis2(d)}]{\includegraphics[width=3.4in]{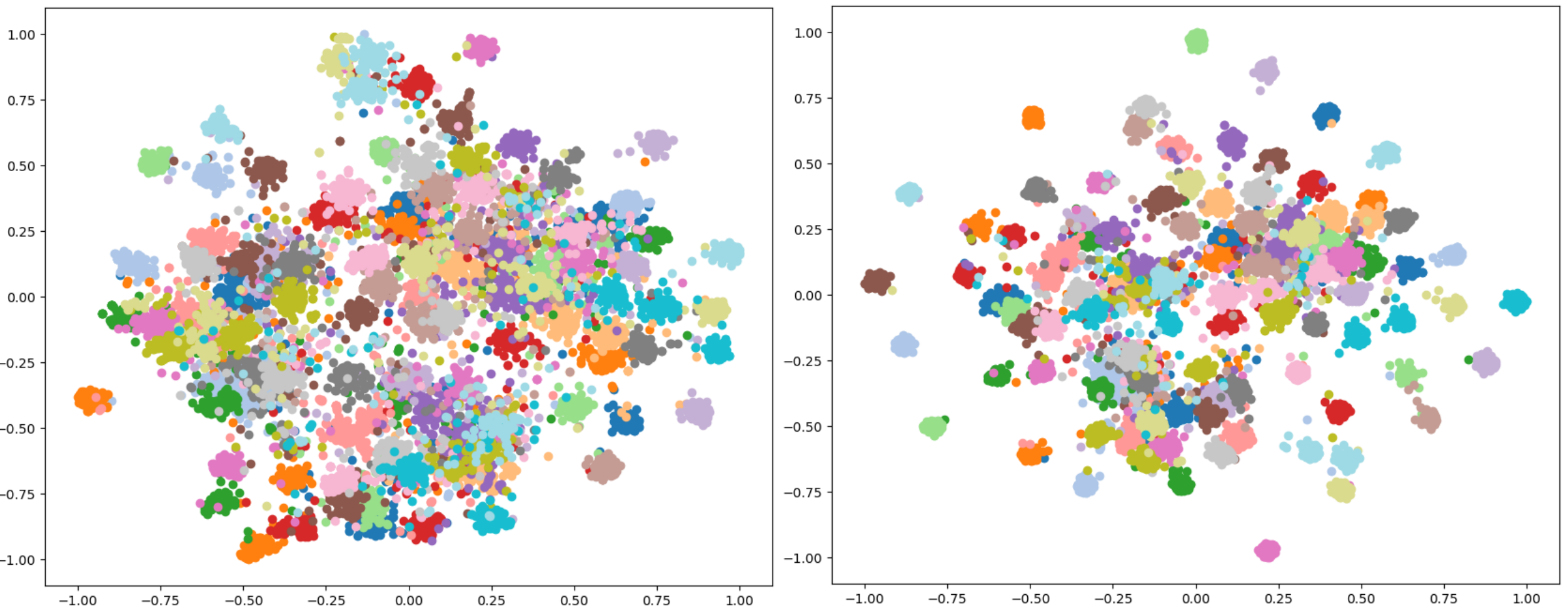}}
\caption{T-SNE visualization of original features and corresponding task subspace projection features. In each group, the left is the visualization of the original features, and the right is the visualization of subspace projection features corresponding to true task.} \label{fig:vis2}
\end{figure*}

\paragraph{Visualizations} We perform t-SNE visualization in the CIFAR-100 dataset. As illustrated in Fig.~\ref{fig:vis1}, the model incrementally learns 5 categories per task, and we provide t-SNE visualizations of the original features and subspace projection features after learning three tasks. As shown in Fig.~\ref{subfig:vis1(a)}, the features for different tasks are dispersed with overlapped boundaries between categories. As illustrated in Fig.~\ref{subfig:vis1(b)},~\ref{subfig:vis1(c)} and~\ref{subfig:vis1(d)}, in each task subspace, in-task samples cluster around their respective class centers, while out-of-task samples remain dispersed, thereby validating our motivation for \textsc{Miles}. Furthermore, we compare the t-SNE visualizations of original features and subspace projection features of corresponding task space in the incremental learning process. As shown in Fig.~\ref{fig:vis2}, subspace projection features exhibit more distinct class boundaries, whereas the original features demonstrate significant boundary ambiguity, providing compelling validation for our research motivation.

\paragraph{Motivation Verification} The motivation diagram in Fig.~\ref{fig:motivation} forms the basis of the proposed method, namely that features of similar samples naturally cluster in the feature space of pre-trained models. To further validate this prior, we conducted a metric analysis on four common pre-trained models in Tab.~\ref{tab:dis_analysis}. Specifically, we calculated the \textbf{average distance from each sample to its class prototype} (denoted as intra-class distance) and the \textbf{average distance from each sample to other class prototypes} (denoted as inter-class distance) on six public benchmark datasets. The results show that in the feature space of the model with \textbf{random} weights, there is a significant confusion interval between intra-class and inter-class distances, within which it is impossible to determine which class a specific sample belongs to. In contrast, in the feature space of \textbf{pre-trained} models, \textbf{the intra-class distance is significantly smaller than the inter-class distance}. Notably, the intra-class distance under random weights is also smaller than the inter-class distance. The reason is that for any given sample, there are always multiple distant class prototypes that raise the average distance. The standard deviation indicates a considerable confusion interval between the two, making classification with random weights infeasible.

\section{conclusion} 
This paper proposes metric learning with expandable subspace~(\textsc{Miles}) for pre-trained model-based class incremental learning. Driven by the observation that features of unseen classes follow regular distributions in the feature space of the pre-trained model, we first design a \textit{subspace prototype projection} strategy to project original features into a task-specific subspace, where the inter-class boundary confusion is reduced and feature divergence across tasks is amplified. To further mitigate the limited fitting capability of adapters in the SPP strategy, we introduce a \textit{subspace representation extension} strategy that effectively enhances the model representational capacity while minimizing computational overhead. Additionally, to address the overfitting in specific task subspaces, we propose \textit{distance regularization}, which balances feature-to-prototype distances across all subspaces through a regularization term. Extensive experiments verify the effectiveness of \textsc{Miles}.

\textbf{Limitation and future work:} Current study mainly focuses on the standard class-incremental setting, where the pre-trained backbone is assumed to provide transferable representations for incoming tasks and the downstream data largely remain within the target semantic scope. This assumption may be restrictive in real-world applications, where incremental streams often contain out-of-distribution (OOD) shifts in style, acquisition conditions, or semantic content. Under such shifts, the prototype structure and task-specific subspaces learned by \textsc{Miles} may become less reliable. Therefore, although \textsc{Miles} achieves strong performance on PTM-based CIL benchmarks, its robustness under OOD data streams has not yet been systematically investigated. A meaningful future direction is to extend \textsc{Miles} toward OOD-aware class-incremental learning by incorporating uncertainty-aware feature or prototype modeling~\cite{liuncertainty}, content-style based augmentation or benign/malign OOD discrimination~\cite{huangharnessing}, and test-time shift correction before subspace projection and prototype matching~\cite{yu2023distribution}, which may improve its applicability to realistic non-stationary environments with both category growth and distribution drift.

\printbibliography
\vspace*{-1.5\baselineskip}
\begin{IEEEbiography}[{\includegraphics[width=1in,height=1.25in,clip,keepaspectratio]{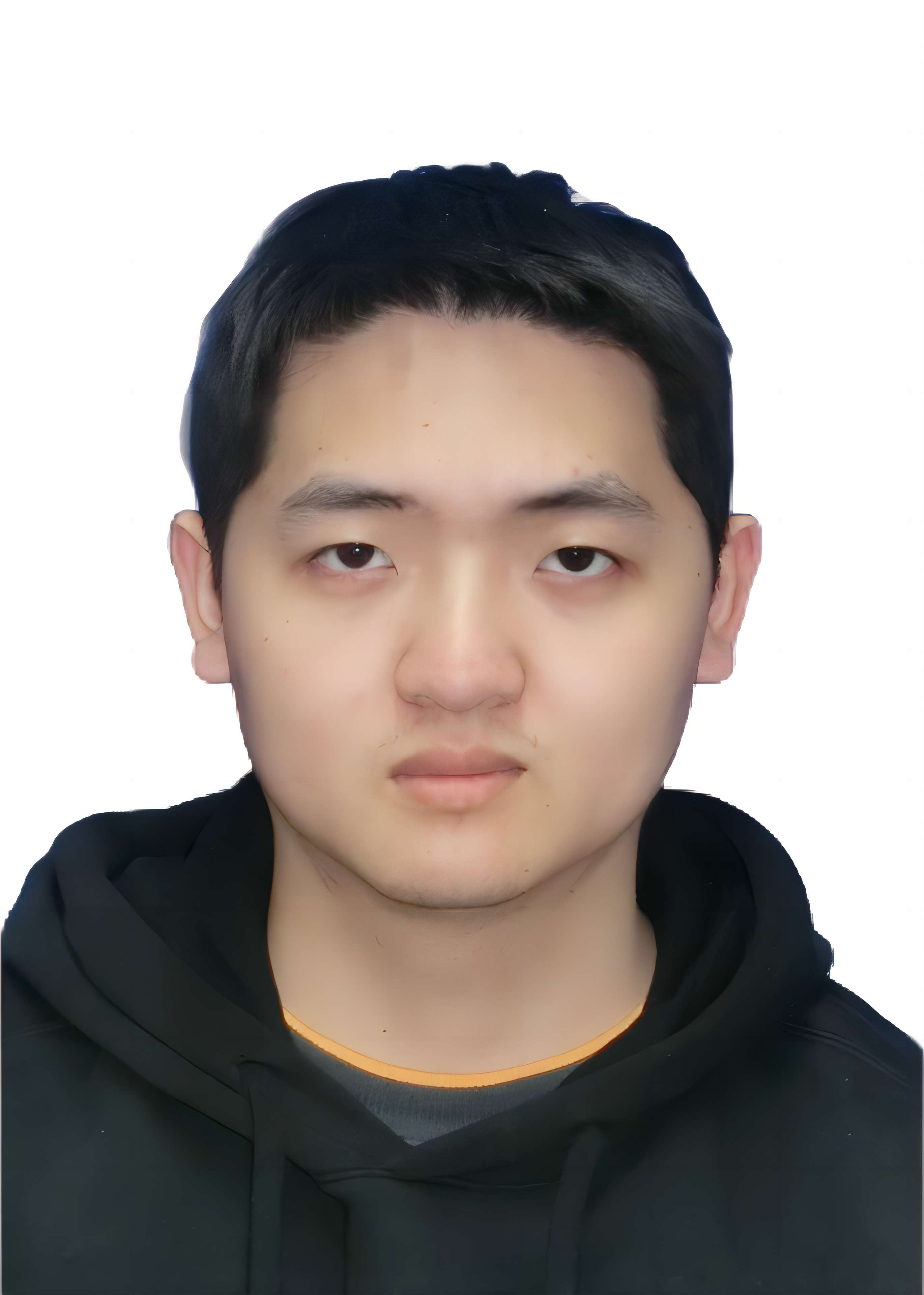}}]{Kai Jiang}
received the B.E. and M.E. degrees from Xidian University, Xi’an, China, in 2022 and 2025, respectively. He is currently pursuing the Ph.D. degree at the School of Artificial Intelligence, Optics and Electronics (iOPEN), Northwestern Polytechnical University, Xi’an, China. His research interests include computer vision and machine learning.
\end{IEEEbiography}
\vspace*{-1.5\baselineskip}
\begin{IEEEbiography}[{\includegraphics[width=1in,height=1.25in,clip,keepaspectratio]{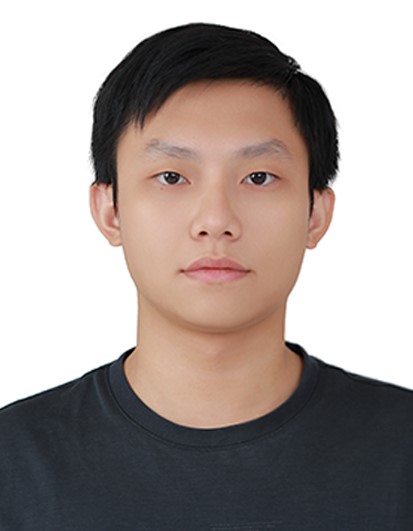}}]{Zisong Lin}
received the B.E. degrees from Xidian University, Xi'an, China, in 2024. He is currently pursuing the M.E degree at the National Key Laboratory of Radar Signal Processing, Xidian University, Xi'an, China. His research interests include computer vision and machine learning.
\end{IEEEbiography}
\vspace*{-1.5\baselineskip}
\begin{IEEEbiography}[{\includegraphics[width=1in,height=1.25in,clip,keepaspectratio]{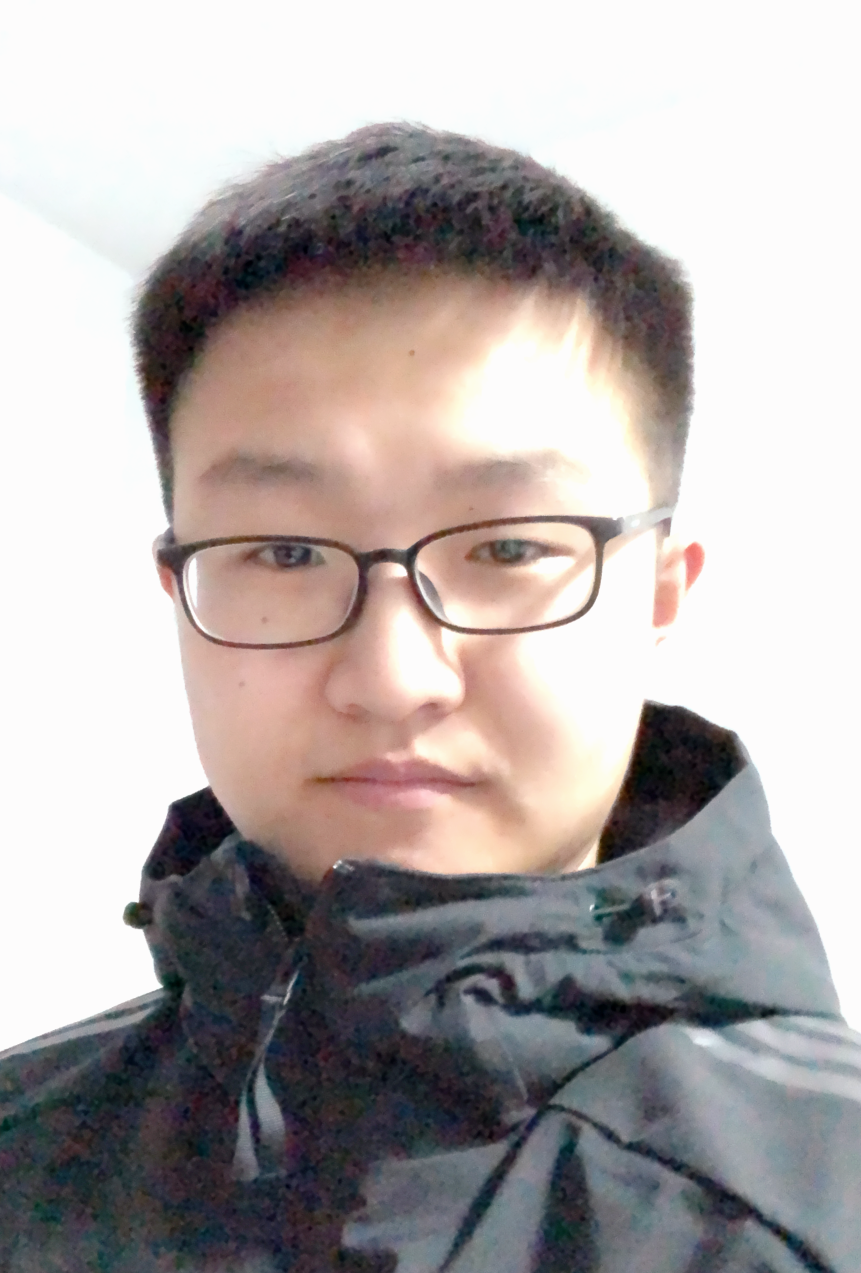}}]{Hongyuan Zhang}
    received the B.E. degree in software engineering from Xidian University, Xi'an, China in 2019 
    and received the Ph.D. degree from the School of Computer Science and the School of Artificial Intelligence, Optics and Electronics (iOPEN), Northwestern Polytechnical University, Xi'an, China in 2024, under the supervision of Prof. Xuelong Li. 
    He is currently a Postdoctoral Fellow of The University of Hong Kong, working with Prof. Ping Luo. 
\end{IEEEbiography}
\vspace*{-1.5\baselineskip}
\begin{IEEEbiography}
[{\includegraphics[
    width=1in,
    height=1.25in,
    clip,
    keepaspectratio
]{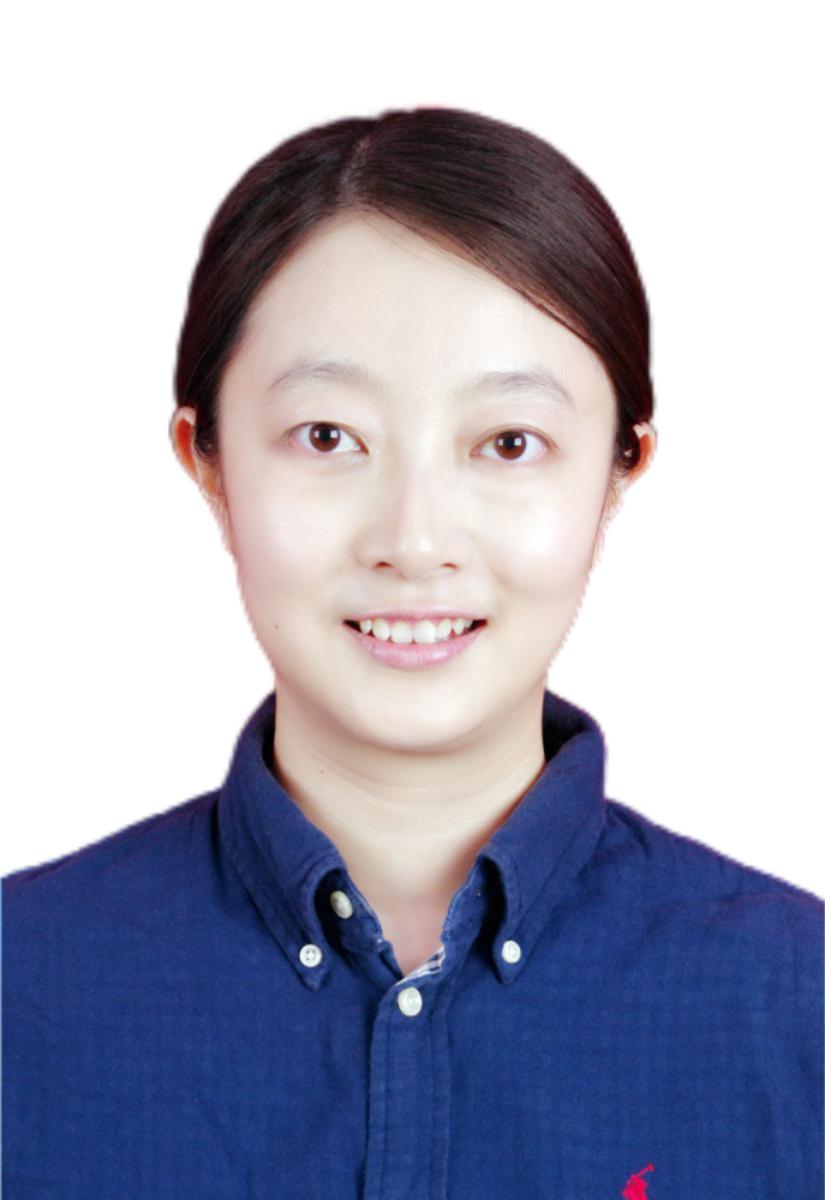}}]
{Xueru Bai}
received the B.S. and Ph.D. degrees in signal and information processing
from Xidian University, Xi'an, China, in 2006 and 2011, respectively.
She is currently a Professor with the National Laboratory of Radar
Signal Processing, Xidian University. Her research interests include
high-resolution radar imaging and radar automatic target recognition.
She was a recipient of the National Excellent Doctoral Dissertation
Award granted by the Ministry of Education of China and the Program
for Excellent Young Scientist selected by the National Natural Science
Foundation of China.
\end{IEEEbiography}
\vspace*{-1.5\baselineskip}
\begin{IEEEbiographynophoto}{Xuelong Li}
(M'02--SM'07--F'12) is the CTO and Chief Scientist of China Telecom,
where he founded the Institute of Artificial Intelligence (TeleAI)
of China Telecom.
\end{IEEEbiographynophoto}
\end{document}